\documentclass{article}

    \PassOptionsToPackage{numbers, compress}{natbib}
\usepackage[preprint]{neurips_2026}

\usepackage[utf8]{inputenc} 
\usepackage[T1]{fontenc}    
\usepackage{hyperref}       
\usepackage{url}            
\usepackage{booktabs}       
\usepackage{amsfonts}       
\usepackage{nicefrac}       
\usepackage{microtype}      
\usepackage{xcolor}         

\usepackage{amsmath}
\usepackage{amssymb}
\usepackage{mathtools}
\usepackage{amsthm}
\usepackage{microtype}
\usepackage{graphicx}
\usepackage{subcaption}
\usepackage{hyperref}

\theoremstyle{plain}

\DeclareMathOperator*{\argmax}{arg\,max}
\usepackage{algorithm}
\usepackage{algorithmic}
\usepackage{enumitem}
\usepackage{multirow}
\usepackage{wrapfig}
\usepackage{caption}
\usepackage{float}

\theoremstyle{plain}

\usepackage{aliascnt}

\theoremstyle{plain}
\newtheorem{theorem}{Theorem}[section]

\newaliascnt{proposition}{theorem}

\aliascntresetthe{proposition}

\newaliascnt{lemma}{theorem}
\newtheorem{lemma}[lemma]{Lemma}
\aliascntresetthe{lemma}

\newaliascnt{corollary}{theorem}

\aliascntresetthe{corollary}

\theoremstyle{definition}

\newaliascnt{definition}{theorem}
\newtheorem{definition}[definition]{Definition}
\aliascntresetthe{definition}

\newaliascnt{assumption}{theorem}

\aliascntresetthe{assumption}

\theoremstyle{remark}

\newaliascnt{remark}{theorem}

\aliascntresetthe{remark}

\usepackage[capitalize,noabbrev,nameinlink]{cleveref}

\crefname{lemma}{lemma}{lemmas}
\Crefname{lemma}{Lemma}{Lemmas}

\crefname{theorem}{theorem}{theorems}
\Crefname{theorem}{Theorem}{Theorems}

\crefname{table}{table}{tables}
\Crefname{table}{Table}{Tables}
\crefname{definition}{definition}{definitions}
\Crefname{definition}{Definition}{Definitions}

\crefname{assumption}{assumption}{assumptions}
\Crefname{assumption}{Assumption}{Assumptions}

\crefname{problem}{problem}{problems}
\Crefname{problem}{Problem}{Problems}

\title{Soft Deterministic Policy Gradient with Gaussian Smoothing}

\author{%
  Hyunjun Na \quad Donghwan Lee\thanks{Corresponding author.} \\
  School of Electrical Engineering, KAIST\\
  291 Daehak-ro, Yuseong-gu, Daejeon 34141, Korea\\
  \texttt{\{nhjun,donghwan\}@kaist.ac.kr} \\
}

\begin{document}

\maketitle

\begin{abstract}
Deterministic policy gradient (DPG) is widely utilized for continuous control; however, it inherently relies on the differentiability of the critic with respect to the action during policy updates. This assumption is  violated in practical control problems involving sparse or discrete rewards, leading to ill-defined policy gradients and unstable learning. To address these challenges, we propose a principled alternative based on a smoothed Bellman equation formulated via Gaussian smoothing. Specifically, we define a novel action-value function based on a smoothed Bellman equation and derive the soft deterministic policy gradient (Soft-DPG). Our formulation eliminates explicit dependence on critic action-gradients and ensures that the gradient remains well-defined even for non-smooth Q-functions. We instantiate this framework into a deep reinforcement learning algorithm, which we call soft deep deterministic policy gradient (Soft DDPG). Empirical evaluations on standard continuous control benchmarks and their discretized-reward variants show that Soft DDPG remains competitive in dense-reward settings and provides clear gains in most discretized-reward environments, where standard DDPG is more sensitive to irregular critic landscapes.
\end{abstract}

\section{Introduction}
Reinforcement learning (RL) has been successfully applied to a wide range of domains, including robotics, autonomous driving, and control systems, by learning optimal policies through interactions with the environment \citep{kiran2021deep, recht2019tour, tang2025deep}. Depending on how the optimal policy is represented and optimized, RL algorithms are commonly categorized into value-based and policy-based approaches. Value-based methods such as Deep Q-Networks (DQN) \citep{mnih2013playing} have demonstrated remarkable success in discrete action spaces \cite{mnih2015human,hessel2018rainbow,van2016deep}, while policy-based methods, particularly policy gradient approaches, have been widely used for continuous control problems \cite{silver2014deterministic,
lillicrap2015continuous,
schulman2015trust,
schulman2017proximal,
haarnoja2018soft}.
Unlike value-based methods, policy-based methods directly parameterize the policy and optimize its parameters via gradient-based methods. This direct optimization makes them particularly suitable for high-dimensional or continuous action spaces \cite{williams1992simple,sutton1999policy}. Early approaches primarily focused on stochastic policies, known as stochastic policy gradient (SPG) methods, where the policy gradient involves an expectation over the state and action distributions. However, such estimators typically suffer from high variance, which can lead to unstable and inefficient learning. To address the computational inefficiency and high variance inherent in SPG, deterministic policy gradient (DPG) was proposed by \citet{silver2014deterministic}. Unlike stochastic policies, which require expectations over both state and action spaces, deterministic policies only require expectations over the state space. This structural difference results in more efficient estimation and reduces variance, as the estimator no longer needs to integrate over a stochastic action distribution. However, deterministic policies suffer from limited exploration capability compared to stochastic policies, as the latter inherently explore the environment through stochasticity. To overcome this limitation, off-policy learning with exploratory behavior policies has been introduced \citep{silver2014deterministic}. This approach facilitates effective exploration while learning a deterministic target policy.

Building upon these advantages, deep RL algorithms such as Deep Deterministic Policy Gradient (DDPG) \cite{lillicrap2015continuous} and Twin Delayed DDPG (TD3) \cite{fujimoto2018addressing} have been developed. These methods have demonstrated strong performance and robustness, not only on standard continuous control benchmarks but also across a wide range of real-world scenarios, including autonomous driving, robotics, and complex industrial control \cite{qiu2019deep, xu2020deep, bouhamed2020autonomous, liu2021ddpg, yu2021multi}.

Despite their success, DPG relies on a critical assumption:
the action-value function must be differentiable with respect to the action.
However, in many practical environments, this assumption is violated.
For example, sparse or discrete reward structures are commonly used in robotics and autonomous driving \cite{kiran2021deep, vecerik2017leveraging}, leading to non-smooth or even non-differentiable Q-functions.
When such non-smoothness is present, DPG may become ill-defined and can result in unstable or degraded performance in practice. To circumvent these limitations, we propose soft deterministic policy gradient (Soft-DPG), a principled framework that relaxes the stringent differentiability assumption of the critic.
Our approach is based on a novel smoothed Bellman equation formulated via Gaussian smoothing (GS) \cite{nesterov2017random}, which induces a well-behaved action-value function.
By leveraging this smoothed Bellman equation, we derive a new policy gradient theorem that eliminates the explicit dependence on the critic's action-gradient.
Consequently, the resulting policy gradient remains well-defined and stable even when the underlying Q-function is non-smooth.

The contributions of this work are as follows:
\begin{enumerate}
\item[(a)] We introduce Soft-DPG, a novel framework designed to overcome the limitations of standard DPG in environments with non-smooth action-value functions. To derive the Soft-DPG, we introduce a smoothed Bellman equation and provide a rigorous theoretical derivation of the Soft-DPG theorem.

\item[(b)] We establish analytical upper bounds for the approximation errors in both the action-value and state-value functions between the original MDP and the one induced by the smoothed Bellman equation. These bounds provide a formal guarantee on the bias introduced by the smoothing parameter.

\item[(c)] We instantiate our theoretical framework into a practical deep RL algorithm named soft deep deterministic policy gradient (Soft DDPG).
Through extensive experiments on standard continuous control benchmark and their discretized reward variants, we empirically validate that Soft DDPG offers competitive stability and performance, particularly in challenging control tasks with irregular reward surfaces.
\end{enumerate}

\section{Related works}

Deterministic policy gradient (DPG) updates the actor using the gradient of the critic with respect to the action.
Unlike stochastic policy gradient (SPG), where the accuracy of the value estimate is the primary concern, DPG critically depends on the accuracy, and even the existence, of action gradients of the critic.
This fundamental dependency has motivated several prior studies. 

For example, \citet{balduzzi2015compatible} argued that, in DPG, the standard critic learned via temporal-difference (TD) learning is not necessarily suitable for policy optimization.
To address this, they proposed an approach that explicitly learns action gradients using an additional deviant network.
While this approach addresses the mismatch between TD learning and gradient estimation, it introduces extra networks and increases the number of learnable parameters.
They also considered zeroth-order gradient estimation techniques, which are conceptually related to our motivation, but at the cost of additional architectural complexity. 
The most closely related works to ours are those by \citet{nachum2018smoothed, kumar2020zeroth,saglam2024compatible}. These works share the common idea of using smoothed value functions as alternatives to the original Q-function, without introducing additional networks and within a model-free framework.
In particular, \citet{nachum2018smoothed} introduced smoothed Q-functions in the context of stochastic policies. Their objective was not to derive DPG, but rather to enable Gaussian policy optimization by extracting gradients of both the mean and covariance from the smoothed value function. \citet{kumar2020zeroth} and \citet{saglam2024compatible} employed smoothed Q-functions as surrogates for the action gradients of the original critic in DPG.
By smoothing the Q-function directly, these approaches reduce sensitivity to local non-smoothness and noisy gradient estimates.
Our formulation differs from these approaches in where smoothing is applied.
Rather than externally smoothing a learned critic or using smoothed values only as surrogate action-gradient estimators, we incorporate Gaussian smoothing (GS) directly into the Bellman backup. A related idea of smoothing has also been explored in \citet{fujimoto2018addressing}; however, their approach applies smoothing at the level of the target as a regularization technique, whereas our method introduces smoothing directly into the Bellman backup, leading to a Bellman-consistent formulation.




\section{Preliminaries}
\subsection{Markov decision problem}
We consider the infinite-horizon discounted Markov decision problem and Markov decision process (MDP), where the agent sequentially takes actions to maximize cumulative discounted rewards. In a MDP with the state-space ${\cal S}:=\{ 1,2,\ldots ,|{\cal S}|\}$ and action-space ${\cal A}:= \{1,2,\ldots,|{\cal A}|\}$, the decision maker selects an action $a \in {\cal A}$ at the current state $s\in {\cal S}$, then the state
transits to the next state $s'\in {\cal S}$ with probability $P(s'|s,a)$, and the transition incurs a
reward $r(s,a,s') \in {\mathbb R}$, where $P(s'|s,a)$ is the state transition probability from the current state
$s\in {\cal S}$ to the next state $s' \in {\cal S}$ under action $a \in {\cal A}$, and $r(s,a,s')$ is the reward function. Moreover, $|{\cal S}|$ and  $|{\cal A}|$ denote cardinalities of $\cal S$ and $\cal A$, respectively. For convenience, we consider a deterministic reward function and simply write $r(s_k,a_k ,s_{k + 1}) =:r_{k+1},k \in \{ 0,1,\ldots \}$. A deterministic policy, $\pi :{\cal S} \to {\cal A}$, maps a state $s \in {\cal S}$ to an action $\pi(s)\in {\cal A}$. The objective of the Markov decision problem is to find an optimal policy, $\pi^*$, such that the cumulative discounted rewards over infinite time horizons is maximized, i.e.,
\begin{align*}
\pi^*:= \argmax_{\pi\in \Theta} {\mathbb E}\left[\left.\sum_{k=0}^\infty {\gamma^k r_{k+1}}\right|\pi\right],
\end{align*}
where $\gamma \in [0,1)$ is the discount factor, $\Theta$ is the set of all deterministic policies, $(s_0,a_0,s_1,a_1,\ldots)$ is a state-action trajectory generated by the Markov chain under policy $\pi$, and ${\mathbb E}[\cdot|\pi]$ is an expectation conditioned on the policy $\pi$. Moreover, for finite state–action pairs $(s,a)\in\mathcal{S}\times\mathcal{A}$, the Q-function under policy $\pi$ is defined as
\[
Q^\pi(s,a)
= \mathbb{E}\!\left[\sum_{k\ge0}\gamma^k r_{k+1}\mid s_0=s,\;a_0=a,\;\pi\right].
\]
and the optimal Q-function is defined as $Q^*(s,a)=Q^{\pi^*}(s,a)$ for all $(s,a)\in {\cal S} \times {\cal A}$. Once $Q^*$ is known, then an optimal policy can be retrieved by the greedy policy $\pi^*(s)=\argmax_{a\in {\cal A}}Q^*(s,a)$. Throughout, we assume that the MDP is ergodic so that the stationary state distribution exists.

\subsection{Deterministic policy gradient}
\label{sec:dpg}
DPG is a class of policy gradient algorithms that optimize deterministic policies in continuous action spaces \citep{silver2014deterministic}.
Unlike SPG, which model the policy as a probability distribution, DPG assumes a deterministic policy of the form
$
a = \pi_\theta(s)
$
which directly maps states to actions. Accordingly, the objective of a deterministic policy can be written as
\begin{equation}
J(\theta)
=
\mathbb{E}_{s \sim d^{\pi_\theta}}
\left[
Q^{\pi_\theta}(s, \pi_\theta(s))
\right],
\end{equation}
where $d^{\pi_\theta}$ denotes the discounted state visitation distribution under policy $\pi_\theta$, defined as
$d^{\pi_\theta}(s') \propto \sum_{t=0}^{\infty} \gamma^t 
\Pr(s_t = s' \mid s_0 \sim p_0, \pi_\theta),
$
that is, the discounted occupancy measure induced by following $\pi_\theta$ from the initial state distribution. 
In contrast to SPG, no expectation over the action space is required.
As a result, the policy gradient admits a simplified form that depends only on an expectation over the state distribution.
\citet{silver2014deterministic} derived the DPG, given by
\begin{equation}\label{eq:dpg}
\nabla_\theta J(\theta)
=
\mathbb{E}_{s \sim d^{\pi_\theta}}
\left[
\nabla_\theta \pi_\theta(s)\,
\nabla_a Q^{\pi_\theta}(s,a)
\big|_{a=\pi_\theta(s)}
\right],
\end{equation}
As a consequence, DPG benefits from reduced gradient variance, often resulting in improved sample efficiency in continuous control tasks. 
However, deterministic policy suffers from limited exploration capability, as they lack inherent stochasticity. 
To mitigate this issue, \citet{silver2014deterministic} showed that DPG can be estimated in an off-policy manner using importance sampling.
This allows learning a deterministic target policy $\pi_\theta$ from data generated by a stochastic behavior policy, thereby decoupling exploration from policy optimization and enabling more efficient exploration in continuous action spaces. 


\subsection{Gaussian smoothing}
\label{sec:gaussian_smoothing}

GS is a fundamental technique in zeroth-order optimization, where gradients are estimated using only function evaluations when direct differentiation is difficult or infeasible \cite{nesterov2017random}.
The core idea is to construct a smoothed surrogate of the original function by perturbing the input with Gaussian noise.
This procedure yields a function that is differentiable everywhere, even when the original function is non-smooth or non-differentiable. We now formally define the Gaussian-smoothed function used throughout this paper.

\begin{definition}
{\cite{nesterov2017random}}
For a given function $f:\mathbb{R}^n \to \mathbb{R}$, the Gaussian-smoothed function $f_\sigma$ is defined as
\begin{align}
f_\sigma(x)
&:= \mathbb{E}_{w \sim \mathcal{N}(0,I)}[f(x + \sigma w)] = \int_{\mathbb{R}^n} f(x + \sigma w)\, \mathcal{N}(w;0,I)\, dw,
\end{align}
where $\sigma > 0$ is the smoothing parameter, and $\mathcal{N}(w;0,I)$ is the probability density function of the $n$-dimensional standard Gaussian distribution. 
\end{definition}

The Gaussian-smoothed function admits a well-defined gradient, which can be expressed without requiring access to the gradient of the original function. The following result gives a gradient representation of the Gaussian-smoothed function that does not require differentiating the original function.

\begin{definition}
{\cite{nesterov2017random}}
For any function $f:\mathbb{R}^n \to \mathbb{R}$, the Gaussian-smoothed function $f_\sigma(x)$ is differentiable for all $x$, and its gradient is given by
\begin{equation}\label{gradient-gs}
\nabla_x f_\sigma(x)
=
\mathbb{E}_{w \sim \mathcal{N}(0,I)}
\left[
\frac{f(x + \sigma w)}{\sigma}\, w
\right].
\end{equation}
\end{definition}

This identity forms the basis of zeroth-order gradient estimators. Moreover, when the original function $f$ is differentiable, the convergence of the smoothed gradient to the true gradient can be formally stated as the following lemma:

\begin{lemma}{\cite{nesterov2017random}}
\label{lem:smoothed_gradient_convergence}
Let $f:\mathbb{R}^n \to \mathbb{R}$ be a differentiable function. As the smoothing parameter $\sigma$ approaches zero, the gradient of the Gaussian-smoothed function $f_\sigma(x)$ converges to the true gradient of the original function $f(x)$, such that:
\begin{equation}
\lim_{\sigma \to 0} \nabla_x f_\sigma(x) = \nabla_x f(x).
\end{equation}
\end{lemma}
Under mild regularity conditions, the Euclidean distance between the gradients of the original function and its smoothed counterpart can be explicitly bounded~\cite{nesterov2017random}.
As a result, optimizing the smoothed function serves as a practical surrogate for optimizing the original objective.
It has also been shown that the maximum of the smoothed function provides a lower bound on the maximum of the original function, often inducing a smoother and more favorable optimization landscape~\cite{salimans2017evolution}.

\subsection{From stochastic to deterministic policy gradient via Gaussian smoothing}
\label{sec:dpg_smoothing_connection}

In this section, we show that DPG can be interpreted as a limiting case of SPG through the lens of GS. 
For completeness and pedagogical clarity, we provide a self-contained derivation that makes this connection explicit. 
Our analysis highlights how DPG naturally arises from smoothed value functions as formalized in the following lemma: 




\begin{lemma}
\label{lem:dpg-gaussian}
Let $\nu_{\theta}(a|s)$ be a Gaussian policy of the form
$
\nu_{\theta}(a|s)
= \mathcal{N}(a;\,\pi_\theta(s),\,\sigma^2 I),
$
where
$
a = \pi_\theta(s) + \sigma w,
w \sim \mathcal{N}(0,I).
$
Let $J(\nu_\theta)$ denote the expected discounted return under $\nu_\theta$. 

Recall the original stochastic policy gradient theorem:
\begin{equation}
\label{eq:spg_original}
\nabla_\theta J(\nu_\theta)
=
\frac{1}{1-\gamma}
\sum_{s} d^{\nu_\theta}(s)
\int_a Q^{\nu_\theta}(s,a)\, \nabla_\theta \nu_\theta(a|s)\, da.
\end{equation}

For this Gaussian policy, the stochastic policy gradient can be equivalently expressed as
\begin{equation}
\label{eq:spg_gaussian}
\nabla_\theta J(\nu_\theta)
=
\frac{1}{1-\gamma}\,
\mathbb{E}_{s \sim d^{\nu_\theta}, w \sim \mathcal{N}(0,I)}
\left[
\nabla_\theta \pi_\theta(s)\,
\frac{Q^{\nu_\theta}(s,\pi_\theta(s)+\sigma w)\, w}{\sigma}
\right].
\end{equation}

Moreover, as $\sigma \to 0$, by \Cref{lem:smoothed_gradient_convergence}, we have
\begin{equation}
\label{eq:dpg_limit}
\lim_{\sigma \downarrow 0}
\nabla_\theta J(\nu_\theta)
=
\frac{1}{1-\gamma}\,
\mathbb{E}_{s \sim d^{\nu_\theta}}
\left[
\nabla_\theta \pi_\theta(s)\,
\nabla_a Q^{\pi_\theta}(s,a)
\big|_{a=\pi_\theta(s)}
\right],
\end{equation}
which recovers the deterministic policy gradient.
\end{lemma}

The proof is in \Cref{app:proof-lem-dpg} of Appendix. A closely related idea already appears implicitly in~\citet{silver2014deterministic}, where DPG is derived as a limiting case of SPG.
By making the role of GS explicit, the above result provides a simpler and more transparent derivation.

\section{Action-gradient dependence of the critic}
\label{sec:action_gradient_problem}

\begin{wrapfigure}{r}{0.45\textwidth}
\vspace{-10pt}
\centering
\begin{subfigure}{\linewidth}
    \centering
    \includegraphics[width=\linewidth]{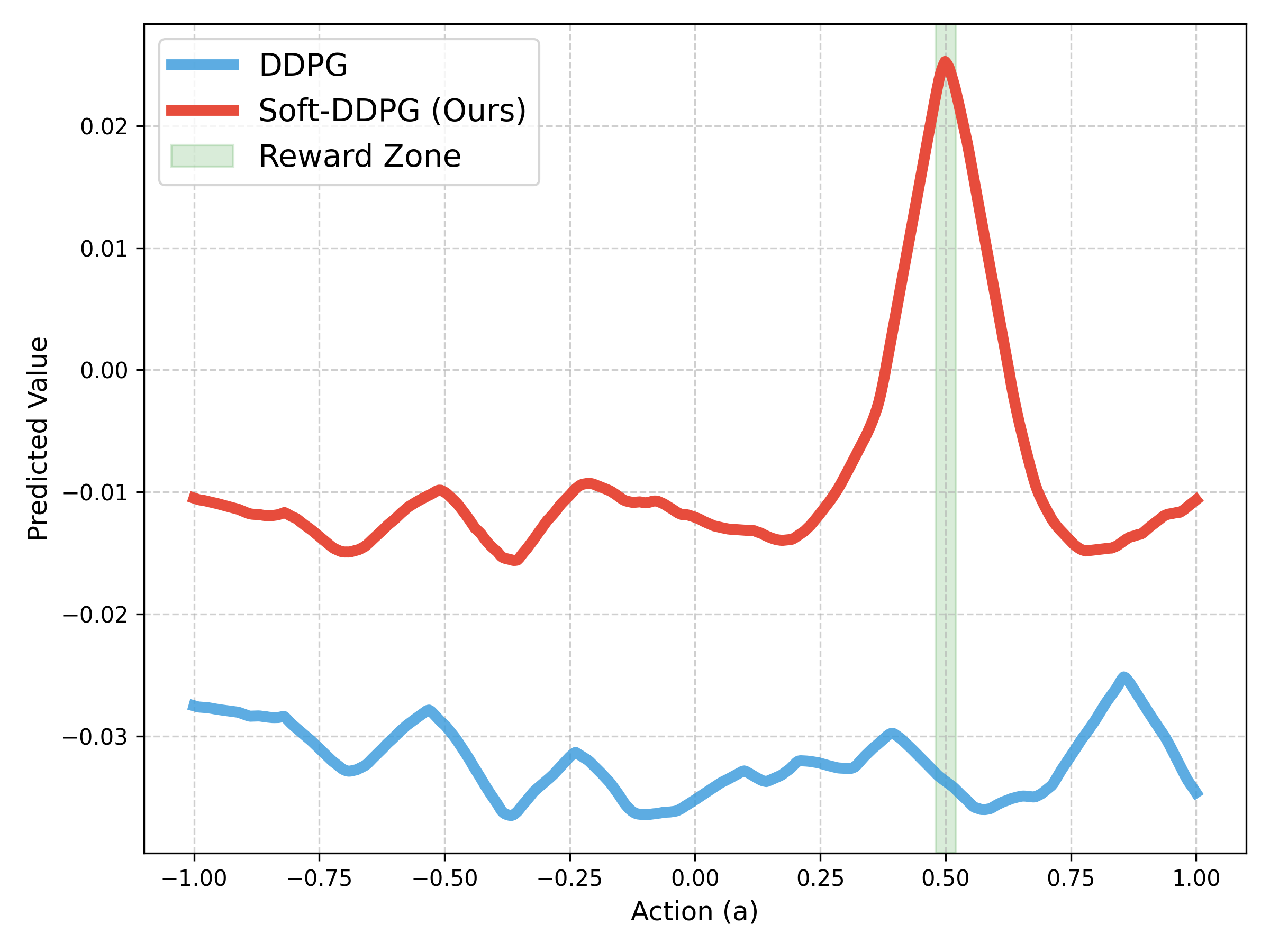}
    \caption{$Q(s,a)$ landscape}
\end{subfigure}
\\[2ex] 
\begin{subfigure}{\linewidth}
    \centering
    \includegraphics[width=\linewidth]{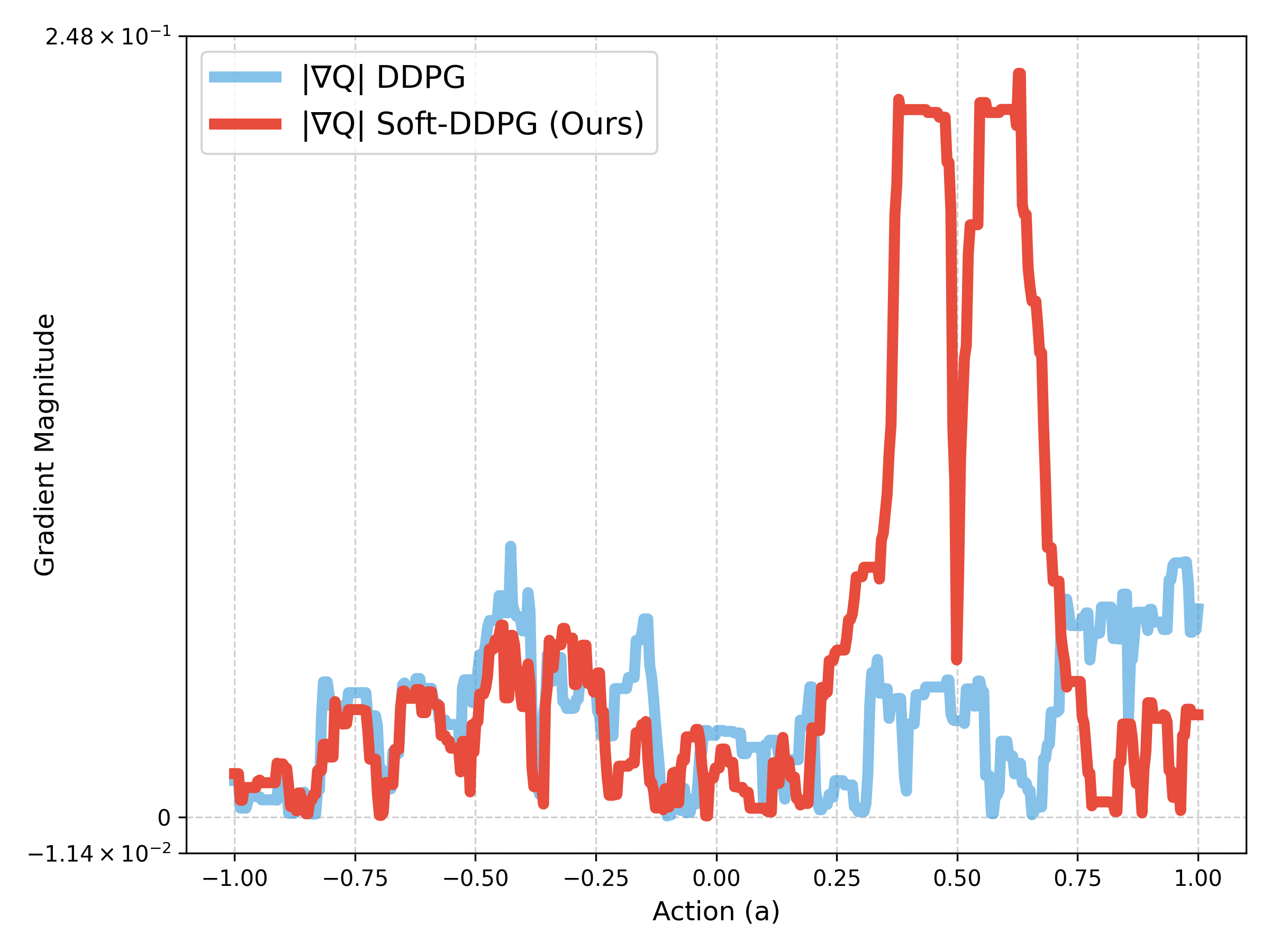}
    \caption{$|\nabla_a Q(s,a)|$ landscape} 
\end{subfigure}
\captionsetup{width=\linewidth}
\caption{
Visualization of the learned critic and its action-gradients in the toy environment with a discrete reward.
}
\label{fig:toy_critic_landscape}
\vspace{-10pt}
\end{wrapfigure}

DPG updates the actor by leveraging the gradient of the critic with respect to the action.
Thus, the policy improvement direction is determined by $\nabla_a Q(s,a)$ evaluated at the current policy action.
This formulation implicitly assumes that the action-gradient of the critic provides a reliable and stable direction for policy improvement. However, in practical implementations of DPG within the actor--critic framework, the critic is trained by minimizing a TD error objective, which enforces accuracy of the value function but does not explicitly constrain the correctness or smoothness of its action-gradient. Particularly in such discrete reward environments, standard TD learning may fail to construct an accurate Q-function landscape, rendering the subsequent action-gradients inherently unreliable. As a result, there is no guarantee that the learned $\nabla_a Q(s,a)$ accurately reflects the true policy improvement direction. 

To illustrate this issue, we construct a simple toy environment designed to expose instability in the critic action-gradient.
The environment consists of a single dummy state $s=[1]$ and a continuous action $a\in[-1,1]$.
The reward function is deliberately discrete: a positive reward is given only when the action lies within a narrow interval around $a=0.5$, and zero otherwise.
Formally, the reward is defined as
\begin{equation*}
r(a) =
\begin{cases}
1, & \text{if } |a-0.5| < \epsilon, \\
0, & \text{otherwise},
\end{cases}
\end{equation*}
where $\epsilon>0$ is a small threshold.
The episode lasts for a fixed number of steps to allow TD learning. To instantiate the DPG framework in practice, we employ the DDPG algorithm to learn the critic, and subsequently analyze both the learned $Q(s,a)$ and its action-gradient $\nabla_a Q(s,a)$. We visualize the learned critic landscape $Q(s,a)$ and the corresponding action-gradient magnitude $|\nabla_a Q(s,a)|$ after training in \Cref{fig:toy_critic_landscape}.
The results show that the critic can exhibit irregular variations across the action space, which directly translate into noisy and unstable gradients.
Since DPG relies directly on these gradients for policy updates, such irregularities can degrade the stability of actor learning. In contrast, as a preview of our proposed method, we also plot the landscape of soft deep deterministic policy gradient (Soft-DDPG) which will be formally introduced in next section. Empirically, we observe that the proposed method yields a more structured critic landscape, which better captures the underlying reward region. Consequently, the induced action-gradients become more stable and informative compared to those of standard DDPG.


\section{Soft deterministic policy gradient}
\label{sec:smoothed_dpg}
To address the limitations of DPG discussed above, we propose a smoothed variant based on GS, which we call soft deterministic policy gradient (Soft-DPG).
Rather than applying smoothing heuristically at the policy or action level,
our approach incorporates GS directly into the Bellman backup operator.
This modification induces a smoothed action-value function that leads to
stable and well-defined policy gradients.


\subsection{$\sigma$-smoothed Bellman equation}
\label{sec:smoothed_bellman}
To address the instability caused by the action-gradient dependence of the critic in DPG,
we introduce a $\sigma$-smoothed Bellman equation, where $\sigma > 0$ is a smoothing parameter.
Our approach applies GS at the level of the Bellman backup operator. This leads to a new smoothed action-value function that induces stable and well-defined policy gradient. Formally, the smoothed action-value function $Q_\sigma^\pi$ is defined as the solution to the following Bellman equation:
\begin{equation}
Q_\sigma^\pi(s,a)
=
R(s,a)
+
\gamma
\int_{\mathcal{S}}
V_\sigma^\pi(s') \, P(s' \mid s,a)\, ds',
\label{eq:smoothed-bellman}
\end{equation}
where the associated smoothed value function is given by
\begin{equation}
V_\sigma^\pi(s)
=
\mathbb{E}_{w \sim \mathcal{N}(0,I)}
\bigl[
Q_\sigma^\pi(s,\,\pi(s)+\sigma w)
\bigr].
\label{eq:smoothed-value}
\end{equation}

Substituting \eqref{eq:smoothed-value} into \eqref{eq:smoothed-bellman} explicitly yields the $\sigma$-smoothed Bellman equation for $Q_\sigma^\pi$:
\begin{equation}
Q_\sigma^\pi(s,a)
=
R(s,a)
+
\gamma
\mathbb{E}_{s' \sim P(\cdot \mid s,a),\, w \sim \mathcal{N}(0,I)}
\bigl[
Q_\sigma^\pi(s',\,\pi(s')+\sigma w)
\bigr].
\label{eq:smoothed-bellman-explicit}
\end{equation}

This relationship motivates the definition of the $\sigma$-smoothed Bellman expectation operator $T_\sigma^\pi$.
\begin{definition}
\label{def:smoothed-operator}
The $\sigma$-smoothed Bellman expectation operator $T_\sigma^\pi$
is defined for any function $Q : \mathcal{S}\times\mathcal{A}\to\mathbb{R}$ as
$
(T_\sigma^\pi Q)(s,a)
:=
R(s,a)
+
\gamma
\mathbb{E}_{s' \sim P(\cdot \mid s,a),\, w \sim \mathcal{N}(0,I)}
\bigl[
Q(s',\,\pi(s')+\sigma w)
\bigr].$
\end{definition}

By comparing \eqref{eq:smoothed-bellman-explicit} and \Cref{def:smoothed-operator}, it directly follows that $Q_\sigma^\pi$ is a fixed point of the operator $T_\sigma^\pi$, satisfying $Q_\sigma^\pi = T_\sigma^\pi Q_\sigma^\pi$.
To guarantee the uniqueness of this fixed point, we must verify that $T_\sigma^\pi$ is a contraction mapping.
The following lemma establishes a key contraction property of the $\sigma$-smoothed Bellman operator:
\begin{lemma}
\label{lemma:contraction-property}
The operator $T_\sigma^\pi$ is a $\gamma$-contraction mapping under the supremum norm.
That is, for any two functions
$Q, Q' : \mathcal{S} \times \mathcal{A} \rightarrow \mathbb{R}$,
\[
\|T_\sigma^\pi Q - T_\sigma^\pi Q'\|_\infty
\le
\gamma \|Q - Q'\|_\infty.
\]
\end{lemma}
The proof is in \Cref{proof:contraction} of Appendix.
Note that it is important to distinguish $Q_\sigma^\pi$ from the GS of the original action-value function $Q^\pi$.
In general,
$
Q_\sigma^\pi(s,a)
\;\neq\;
\mathbb{E}_{w \sim \mathcal{N}(0,I)}
\bigl[
Q^\pi(s, a + \sigma w)
\bigr]$. 
This distinction is crucial for preserving Bellman consistency and enables a principled derivation of Soft-DPG. 
We next quantify the approximation error introduced by GS. First, we establish the error bound for the state-value function through the following lemma: 
\begin{lemma}
\label{lem:smoothing-v-error}
Let $V^\pi$ be the value function of the original MDP and $V_\sigma^\pi$
the value function induced by the $\sigma$-smoothed Bellman equation.
Assume that the reward function $R(s,a)$ is $L_R$-Lipschitz continuous and
the transition density $P(s' \mid s,a)$ is $L_P$-Lipschitz continuous with respect to $a$.
Then
\[
\|V^\pi - V_\sigma^\pi\|_\infty
\le
\frac{\sigma \sqrt{m}}{1-\gamma}
\left(
L_R
+
\frac{\gamma}{2} L_P V_{\max}
\right),
\]
where $m$ is the dimensionality of the action space, and $V_{\max} = R_{\max}/(1-\gamma)$.
\end{lemma}

The proof is in \Cref{proof:bounding_V_error} of Appendix. This result indicates that the bias introduced by GS on the overall value landscape is bounded and can be controlled by the smoothing parameter $\sigma$.  Building upon the Lipschitz continuity of the reward and transition dynamics, we next show that the smoothed action-value function is Lipschitz continuous with respect to the action in the following lemma:

\begin{lemma}
\label{lem:q-lipschitz}
Under the same assumptions as \Cref{lem:smoothing-v-error}, for all $s \in \mathcal{S}$, the smoothed action-value function $Q_\sigma^\pi(s,\cdot)$ is $L_Q$-Lipschitz continuous with respect to the Euclidean norm on the action space, where the Lipschitz constant is given by 
$
L_Q = L_R + \gamma L_P V_{\max}.
$
\end{lemma}
The proof is provided in \Cref{proof:q-lipschitz} of Appendix. We can now formally bound the approximation error between the original action-value function $Q^\pi$ and its smoothed counterpart $Q_\sigma^\pi$.

\begin{lemma}
\label{lem:smoothing-error}
Under the assumptions of \Cref{lem:smoothing-v-error}, the approximation error between the original and smoothed action-value functions satisfies
\[
\| Q^\pi - Q_\sigma^\pi \|_\infty
\;\le\;
\frac{\gamma L_Q \sigma \sqrt{m}}{1-\gamma},
\]
where $m$ is the dimensionality of the action space.
\end{lemma}
The proof is in \Cref{proof:bounding_Q_error} of Appendix. The above results show that the smoothed action-value function $Q_\sigma^\pi$
approximates
the original action-value function with a controllable bias determined by the
smoothing parameter $\sigma$. 

\subsection{Theoretical formulation of Soft-DPG}
\label{sec:smoothed_dpg}

In this section, we explicitly derive the Soft-DPG.
Building upon the $\sigma$-smoothed Bellman equation introduced in the previous section,
we show how GS at the Bellman operator level leads to a well-defined and stable
policy gradient that does not rely on fragile action-gradients of the critic. We begin by deriving a gradient Bellman equation for the smoothed value function $V_\sigma^\pi$,
which serves as a key intermediate result. The following lemma presents this result:

\begin{lemma}
\label{lemma:gradient-Bellman-eq}
Let $V_\sigma^{\pi_\theta}$ be the value function induced by the $\sigma$-smoothed Bellman equation.
Then its gradient satisfies
\begin{align*}
\nabla_\theta V_\sigma^{\pi_\theta}(s)
&= \gamma
\int_{\mathcal{S}}
\nabla_\theta V_\sigma^{\pi_\theta}(s')\,
P_\sigma(s' \mid s,\pi_\theta(s))\, ds'  
+ \nabla_\theta
\int_{\mathbb{R}^m}
Q_\sigma^{\pi_\theta}\!\left(
s,\pi_\theta(s)+\sigma w
\right)
\phi(w)\, dw \Big|_{\pi = \pi_\theta},
\end{align*}
where $\phi(w)$ denotes the standard Gaussian density.
\end{lemma}
The proof is in \Cref{proof:gradient_flow} of Appendix.
\Cref{lemma:gradient-Bellman-eq} shows that the gradient of the smoothed value function
is governed by a Bellman-type recursion, with the policy appearing only through
Gaussian-perturbed actions.
This structure plays a crucial role in eliminating the direct dependence on
the action-gradient of the critic.
Building on this gradient Bellman equation, we now present the main result of this section, which provides an explicit form of the Soft-DPG.
\begin{theorem}
\label{thm:smoothed-dpg-thm}
Let the smoothed action be
$\tilde{a} = \pi_\theta(s) + \sigma w$, where $w \sim \mathcal{N}(0,I)$,
and define the GS distribution
$
\nu_\theta(\tilde{a} \mid s)
=
\mathcal{N}\!\left(
\tilde{a};\, \pi_\theta(s),\, \sigma^{2} I
\right).
$
Then the Soft-DPG is given by
\begin{align*}
\nabla_\theta J_\sigma
&=\mathbb{E}_{s \sim \rho^{\nu_\theta}, \tilde{a} \sim \nu_\theta}\!\left[
-\frac{1}{2\sigma^2}
\nabla_\theta
\bigl\|
\tilde{a}-\pi_\theta(s)
\bigr\|_2^2\,
Q^{\pi_\theta}_\sigma(s,\tilde{a})
\right] 
\\
&=\mathbb{E}_{s \sim \rho^{\nu_\theta}, \tilde{a} \sim \nu_\theta}\!\left[
\frac{1}{\sigma^2}
\nabla_\theta \pi_\theta(s)\,
(\tilde{a}-\pi_\theta(s))\,
Q^{\pi_\theta}_\sigma(s,\tilde{a})
\right],
\end{align*}
where $\rho^{\nu_{\theta}}$ is the discounted state visitation distribution induced by $\nu_{\theta}$.
\end{theorem}
The proof is provided in \Cref{proof:SDPG} of the Appendix, alongside a simpler alternative derivation based on the standard policy gradient in \Cref{proof:alt-sdpg}. \Cref{thm:smoothed-dpg-thm} reveals a key distinction from the classical DPG.
While standard DPG requires the critic $Q(s,a)$ to be differentiable with respect to the action, as shown in \eqref{eq:dpg}, the proposed Soft-DPG depends only on function evaluations of the smoothed critic at perturbed actions, and is therefore well-defined even when the original action-value function is non-smooth. Moreover, unlike prior work that applies GS directly to the original Q-function \cite{nachum2018smoothed,kumar2020zeroth,saglam2024compatible}, our approach uses the smoothed action-value function
defined as the fixed point of a smoothed Bellman operator.
This distinction enables a principled application of smoothing during critic
learning.

\subsection{Soft deep deterministic policy gradient}
\label{sec:smooth_ddpg}

To demonstrate the practical applicability of the proposed Soft-DPG,
we instantiate our framework within a deep RL algorithm.
\Cref{alg:sdpg-q} summarizes the proposed method.
The overall structure closely follows that of standard DDPG, with two key modifications
corresponding to the critic and actor updates. For the critic update, the target value is computed using a Gaussian-perturbed target action,
$
\tilde{a}' = \pi_{\bar{\theta}}(s') + \sigma w,
$ and $ w \!\sim\! \mathcal{N}(0,I),
$
resulting in a smoothed target value. Note that a similar target-smoothing idea was introduced in \citet{fujimoto2018addressing} as a heuristic regularization technique to stabilize critic learning.
However, in our framework, this target smoothing arises directly from the $\sigma$-smoothed Bellman equation that defines the smoothed
action-value function $Q_\sigma^\pi$.
Consequently, the critic is trained to approximate the fixed point of the smoothed Bellman operator,
ensuring consistency with the theoretical formulation.
For the actor update,
instead of relying on the action-gradient of the critic,
the policy is updated by sampling Gaussian-perturbed actions.
This update depends only on evaluations of the smoothed critic and avoids explicit differentiation
of the action-value function with respect to the action, thereby mitigating instability caused by non-smooth critics.

\begin{algorithm}[t]
\caption{Soft DDPG}
\label{alg:sdpg-q}
\begin{algorithmic}[1]
\STATE Initialize $Q_w, \pi_\theta$, targets $\bar w,\bar\theta$, buffer $\mathcal{D}$
\FOR{$t=1$ to $T$}
    \STATE $a = \pi_\theta(s) + \epsilon,\;\epsilon \sim \mathcal{N}(0,\sigma_{expl}^2 I)$ and observe $(r,s')$, store in $\mathcal{D}$, sample $B$
    \STATE Sample $w_i \sim \mathcal{N}(0,I)$ and compute target action $\tilde{a}' = \mathrm{clip}(\pi_{\bar\theta}(s') + \sigma w_i, a_{\min}, a_{\max})$
    \STATE Compute target value $y_i = r + \gamma Q_{\bar w}(s', \tilde{a}')$
    \STATE Update critic parameter by minimizing:
    $\frac{1}{2|B|N}\sum_{i=1}^N \sum_{(s,a,r,s') \in B} (y_i - Q_w(s,a))^2$
    \STATE Sample $w_i \sim \mathcal{N}(0,I)$ and compute perturbed action $a_i = \mathrm{clip}(\pi_\theta(s) + \sigma w_i, a_{\min}, a_{\max})$
    \STATE Update actor parameter by minimizing:
    $\frac{1}{2|B|N\sigma^2}\sum_{i=1}^N \sum_{(s,a,r,s') \in B} \|a_i - \pi_\theta(s)\|^2 Q_w(s,a_i)$
    \STATE Update targets:
    $\bar w \leftarrow (1-\tau) \bar w + \tau w,\;
    \bar\theta \leftarrow (1-\tau) \bar\theta + \tau\theta$
\ENDFOR
\end{algorithmic}
\end{algorithm}

\section{Experiments}
We evaluate the proposed Soft DDPG algorithm on standard MuJoCo \citep{todorov2012mujoco} continuous control benchmarks available in OpenAI Gym \citep{brockman2016openai} and their discrete-reward variants to assess its performance and robustness under different reward structures. Specifically, we compare Soft DDPG with the standard DDPG baseline to isolate the effect of smoothing the DPG. For Soft DDPG, the smoothing parameters, including the smoothing scale $\sigma$ and the number of Monte Carlo samples $N$, are selected based on the sensitivity analysis detailed in \Cref{sec:sensitivity} of Appendix. All other hyperparameters are kept identical for a fair comparison (see \Cref{sec:experiment_detail} of Appendix for detailed configurations).

\begin{figure*}[t]
\centering

\subfloat[Ant (Cont.)]{
\includegraphics[width=0.24\textwidth]{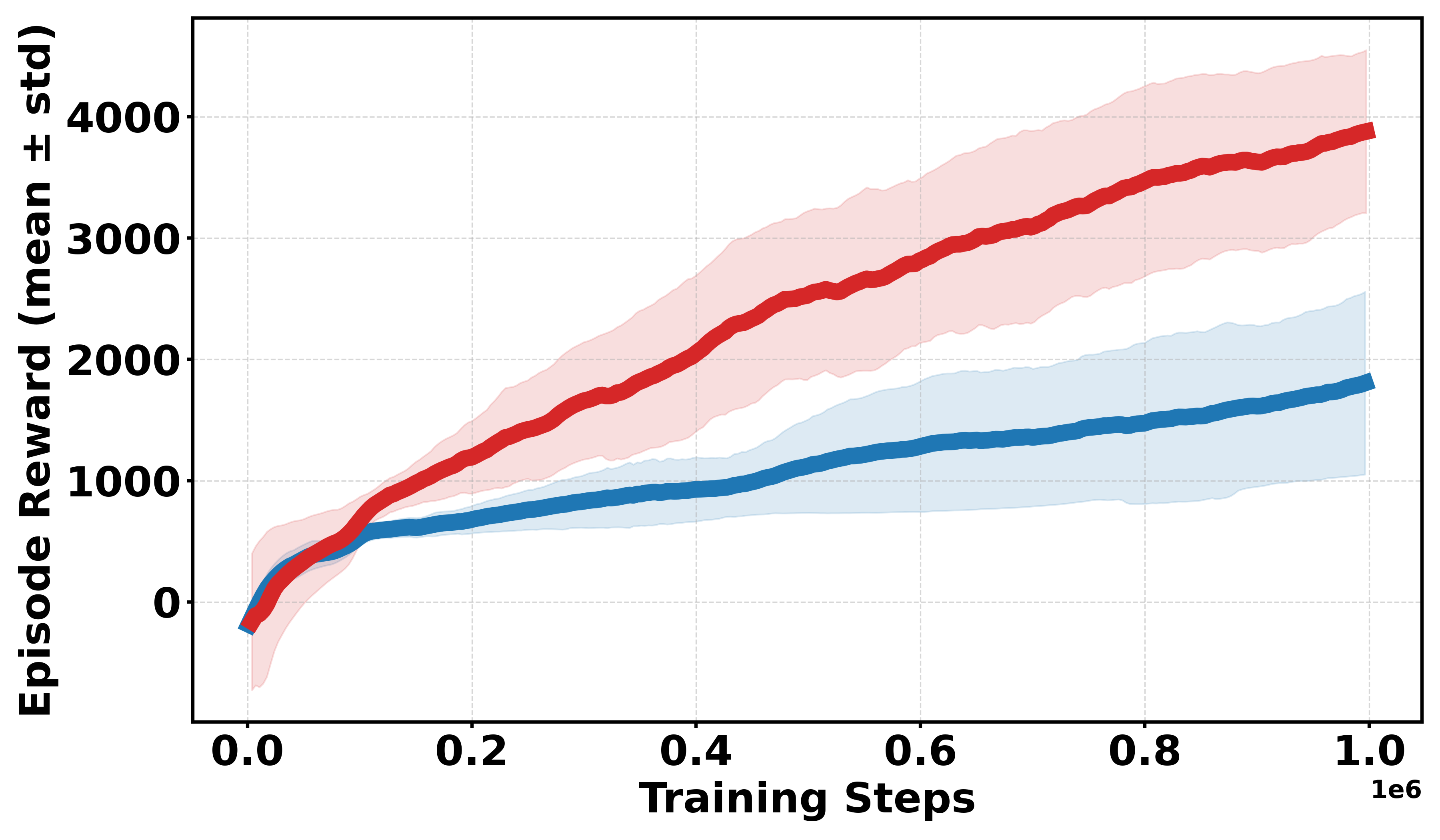}}
\subfloat[HalfCheetah (Cont.)]{
\includegraphics[width=0.24\textwidth]{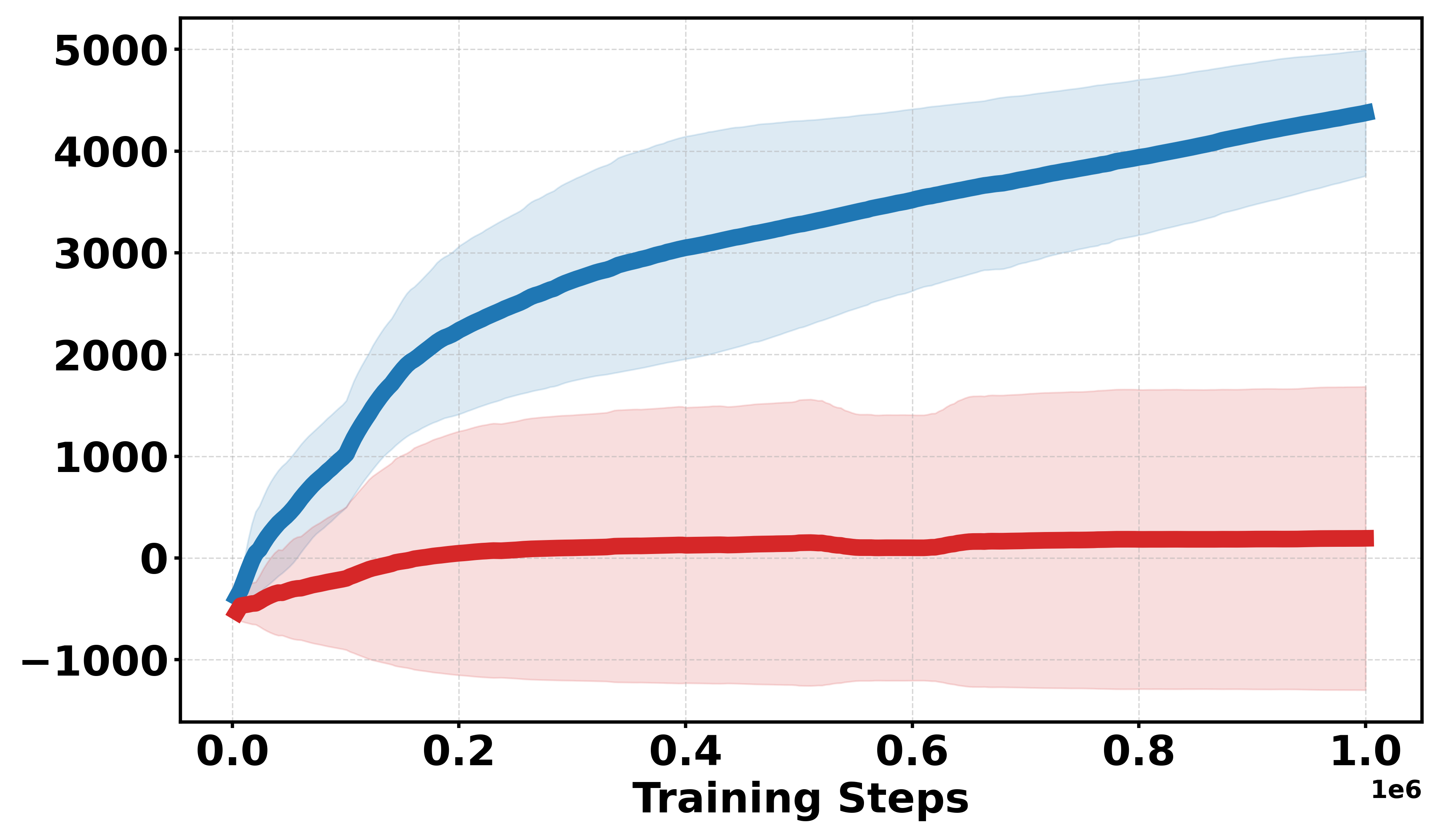}}
\subfloat[Hopper (Cont.)]{
\includegraphics[width=0.24\textwidth]{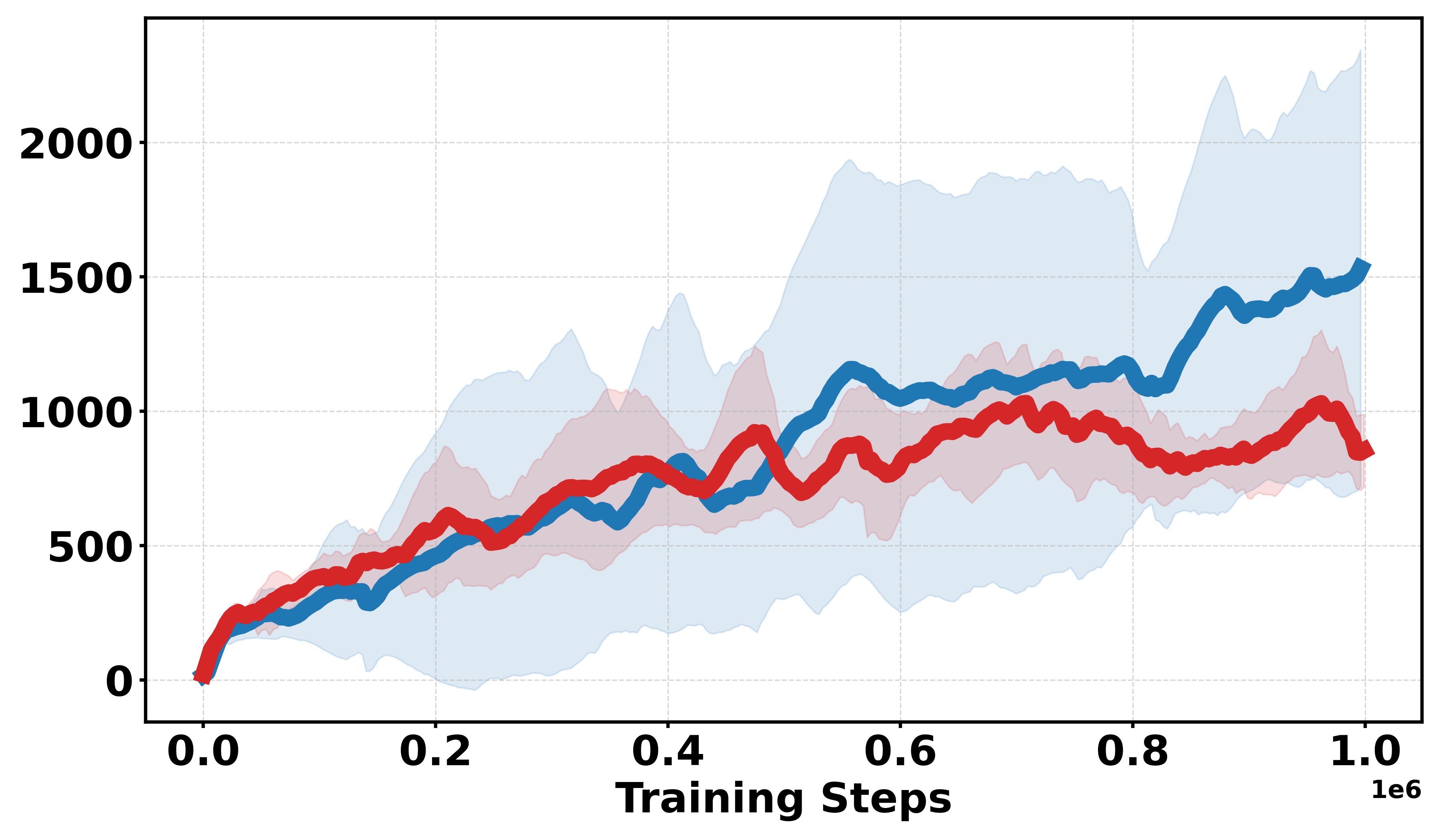}}
\subfloat[Walker2d (Cont.)]{
\includegraphics[width=0.24\textwidth]{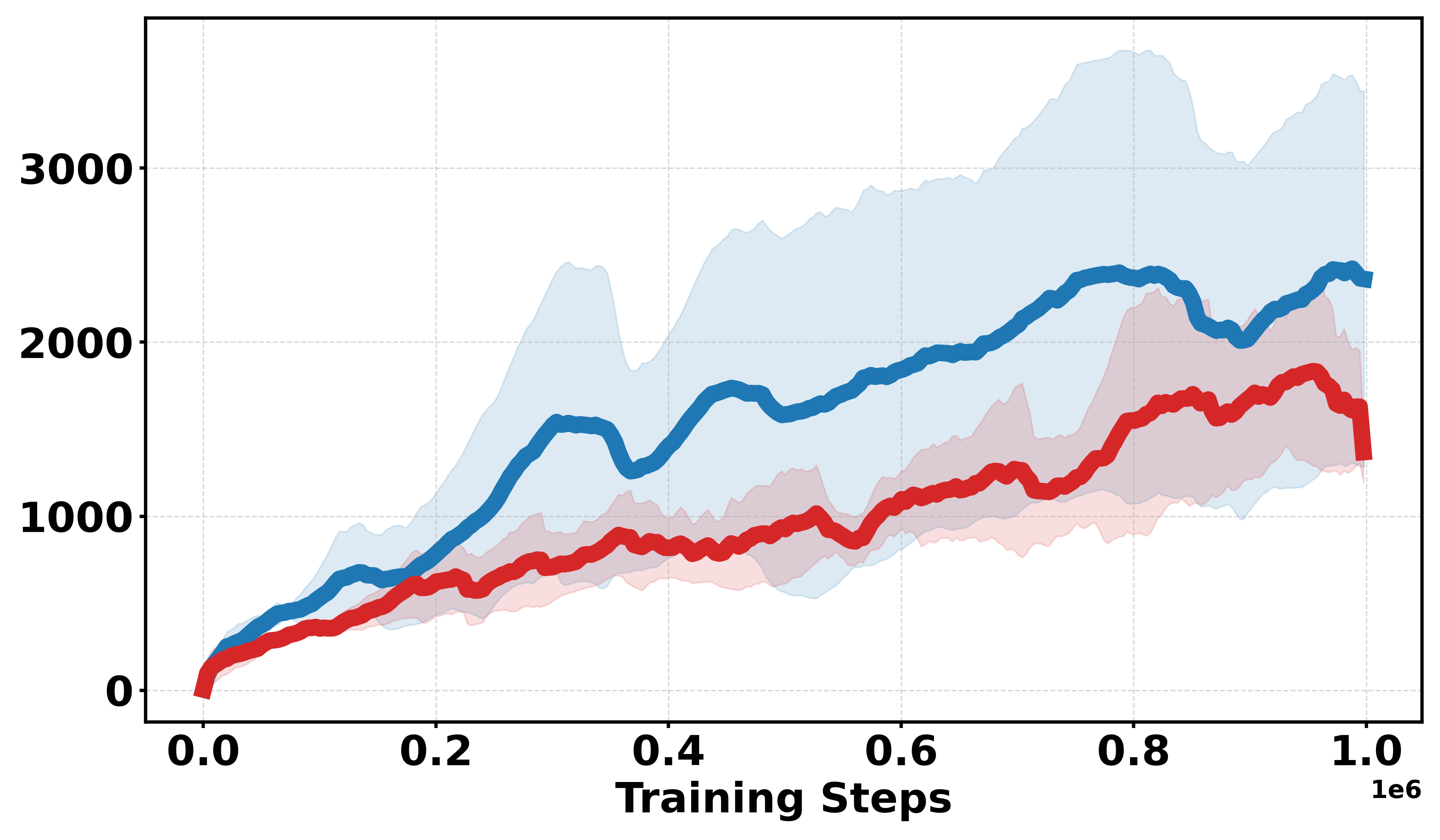}}

\vspace{0.25cm}

\subfloat[Humanoid (Cont.)]{
\includegraphics[width=0.24\textwidth]{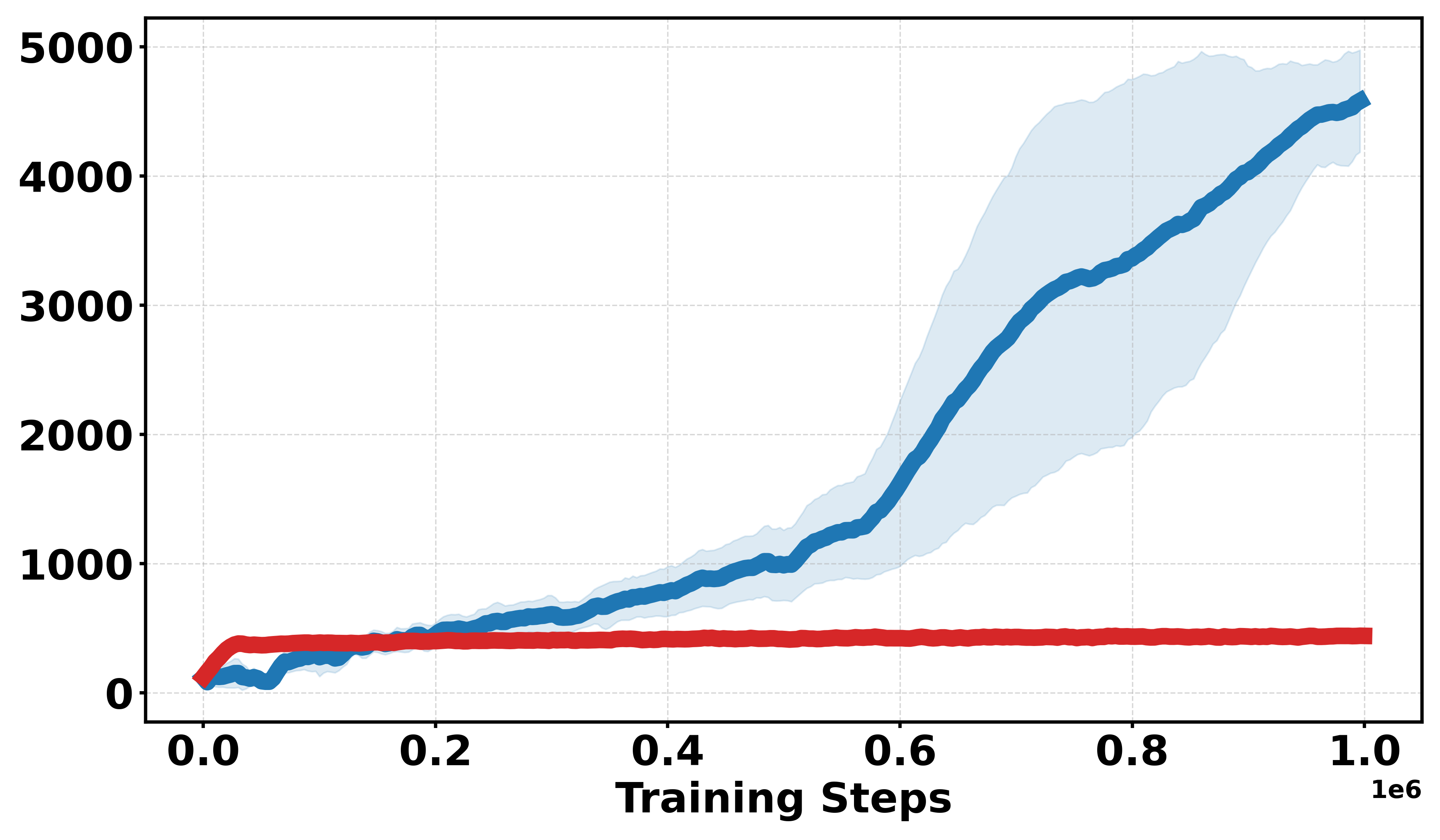}}
\subfloat[Inverted Double Pendulum (Cont.)]{
\includegraphics[width=0.24\textwidth]{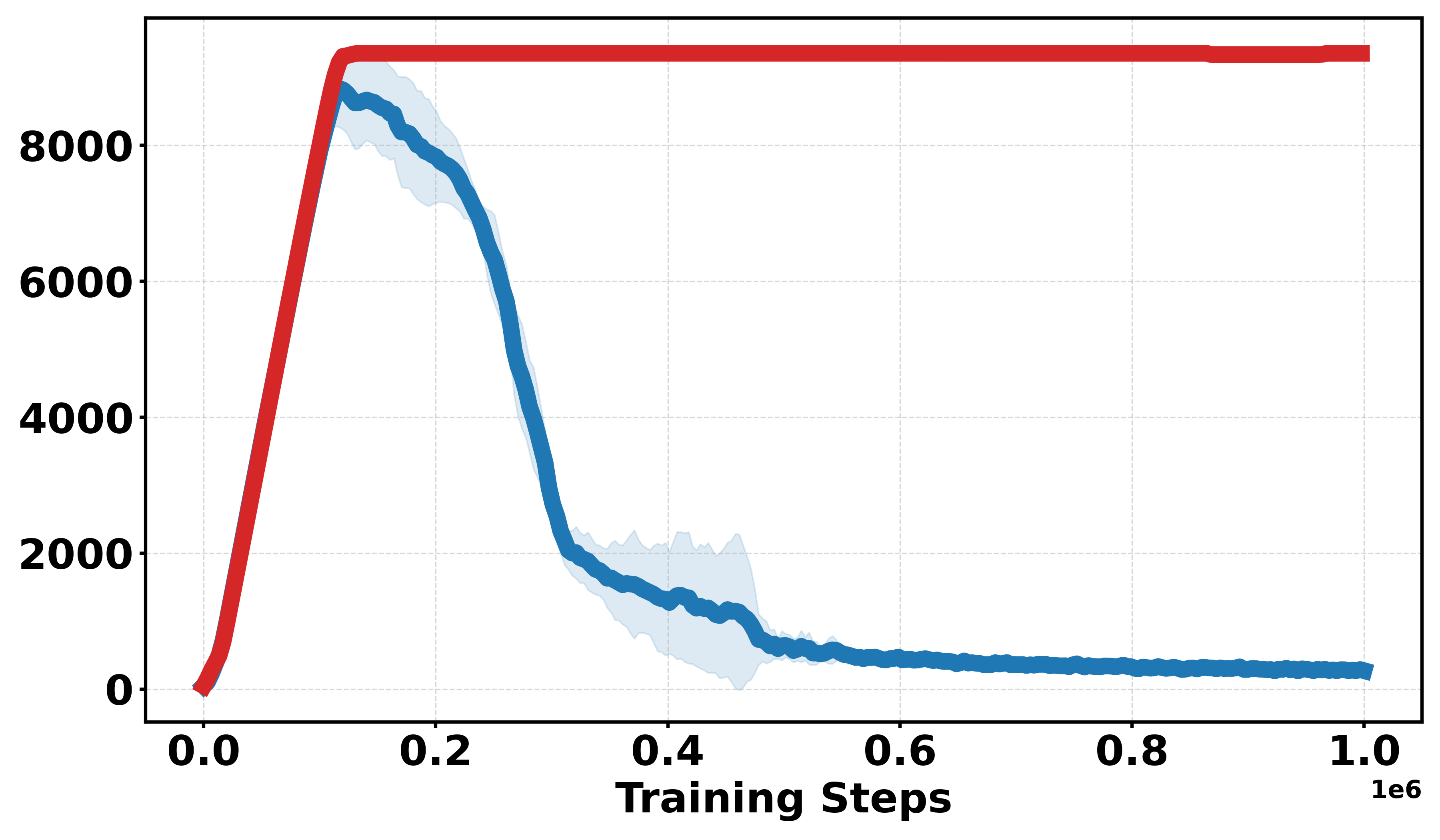}}
\subfloat[Inverted Pendulum (Cont.)]{
\includegraphics[width=0.24\textwidth]{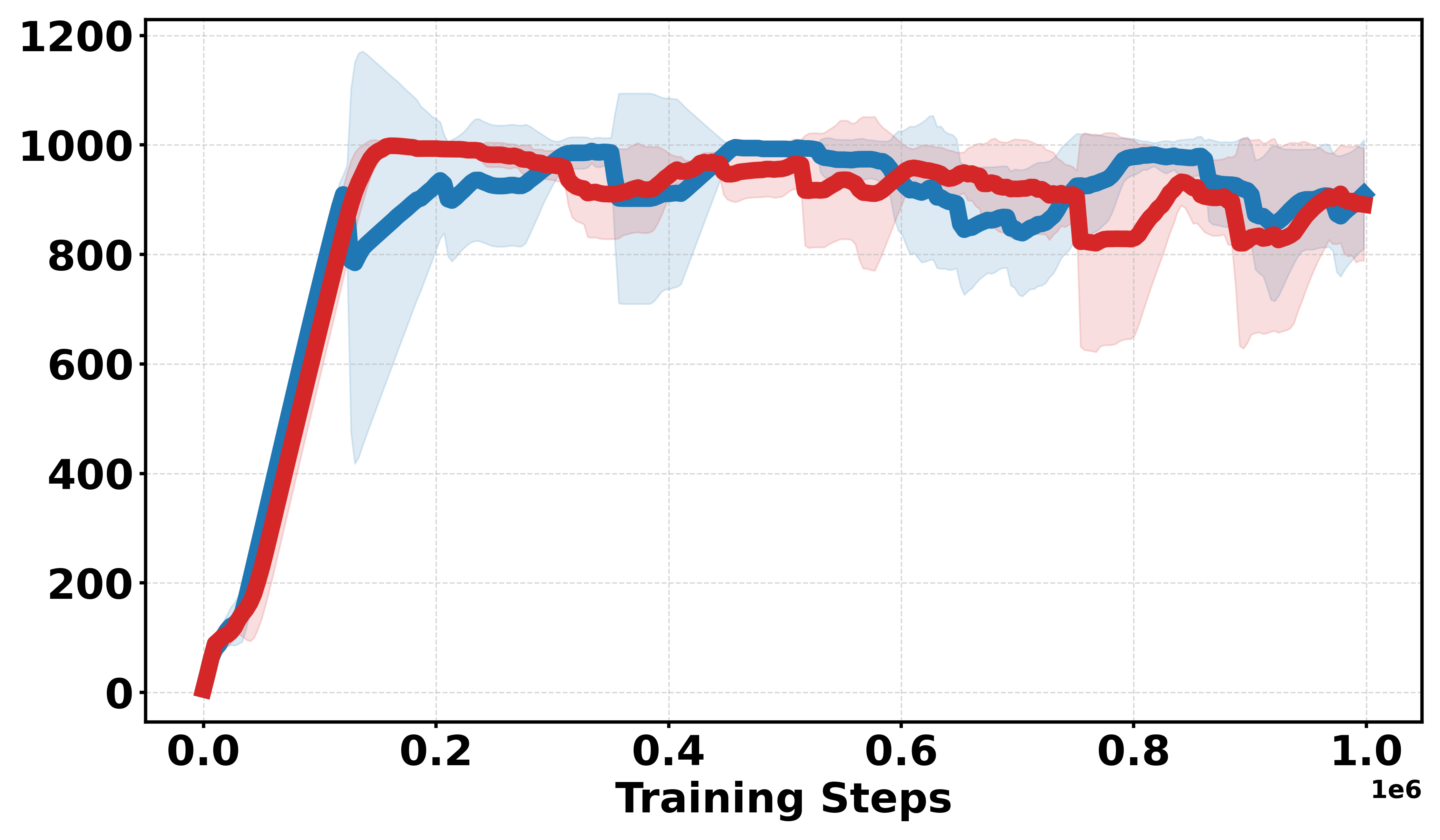}}

\vspace{0.4cm}

\subfloat[Ant (Disc.)]{
\includegraphics[width=0.24\textwidth]{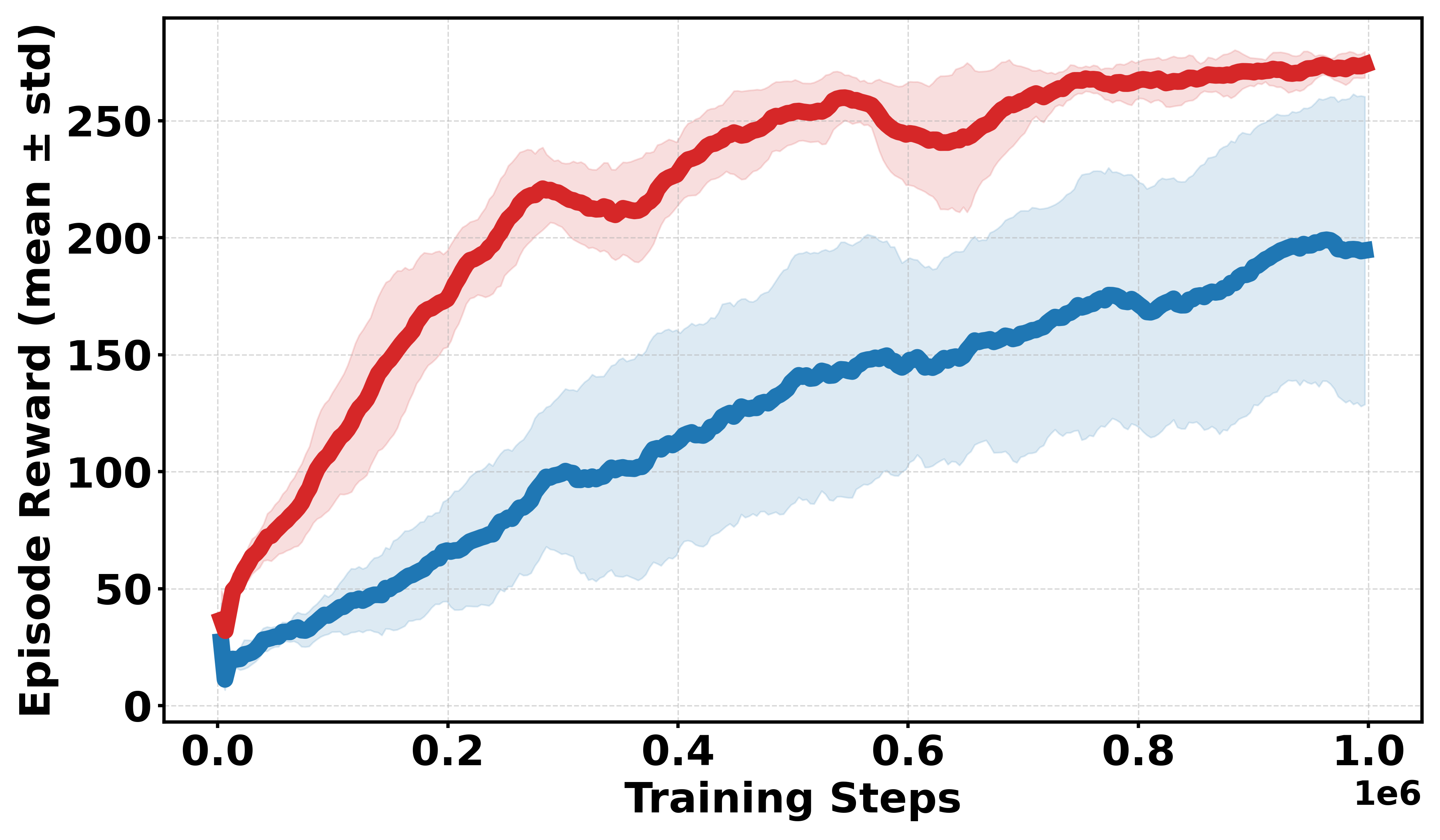}}
\subfloat[HalfCheetah (Disc.)]{
\includegraphics[width=0.24\textwidth]{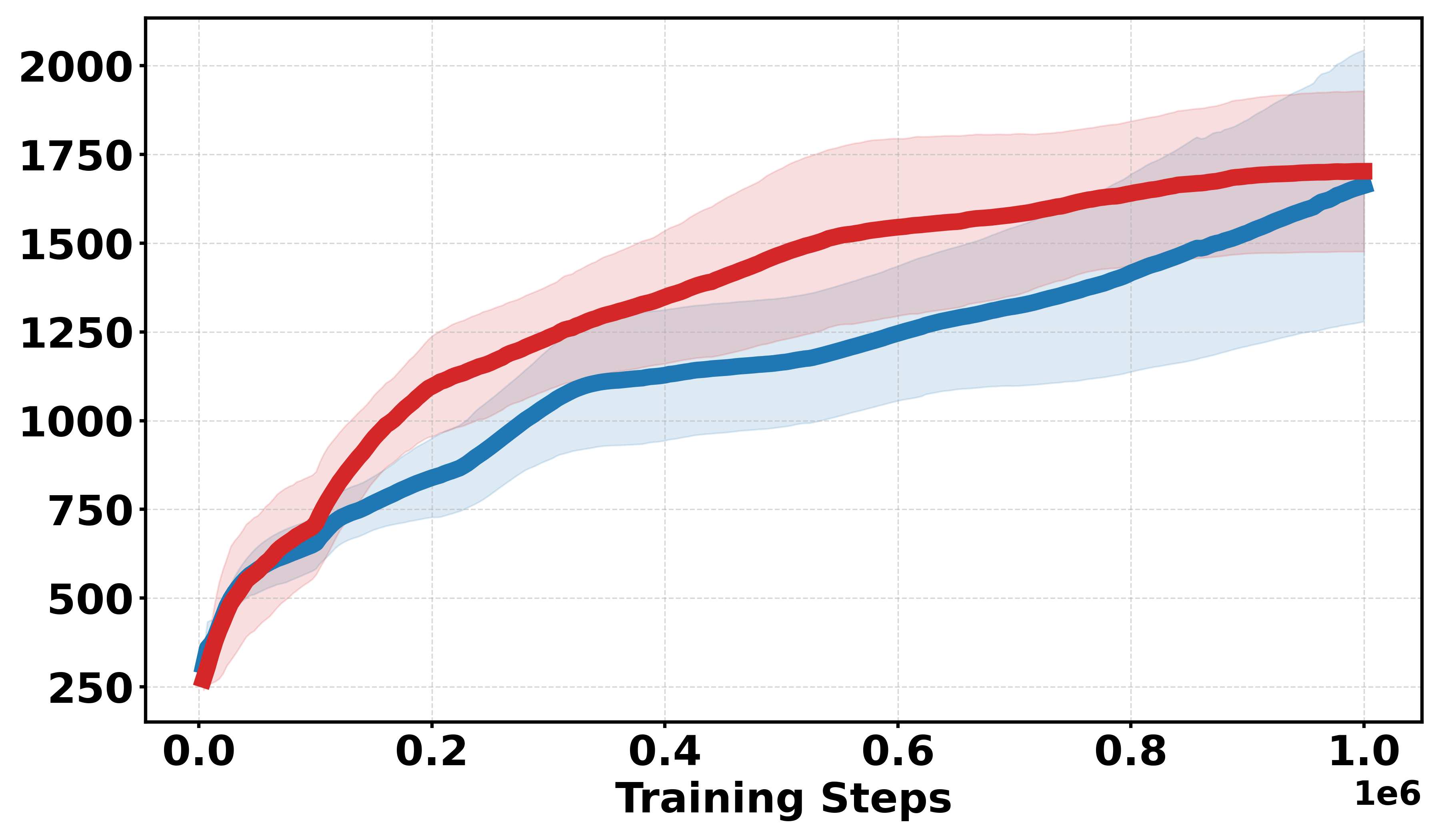}}
\subfloat[Hopper (Disc.)]{
\includegraphics[width=0.24\textwidth]{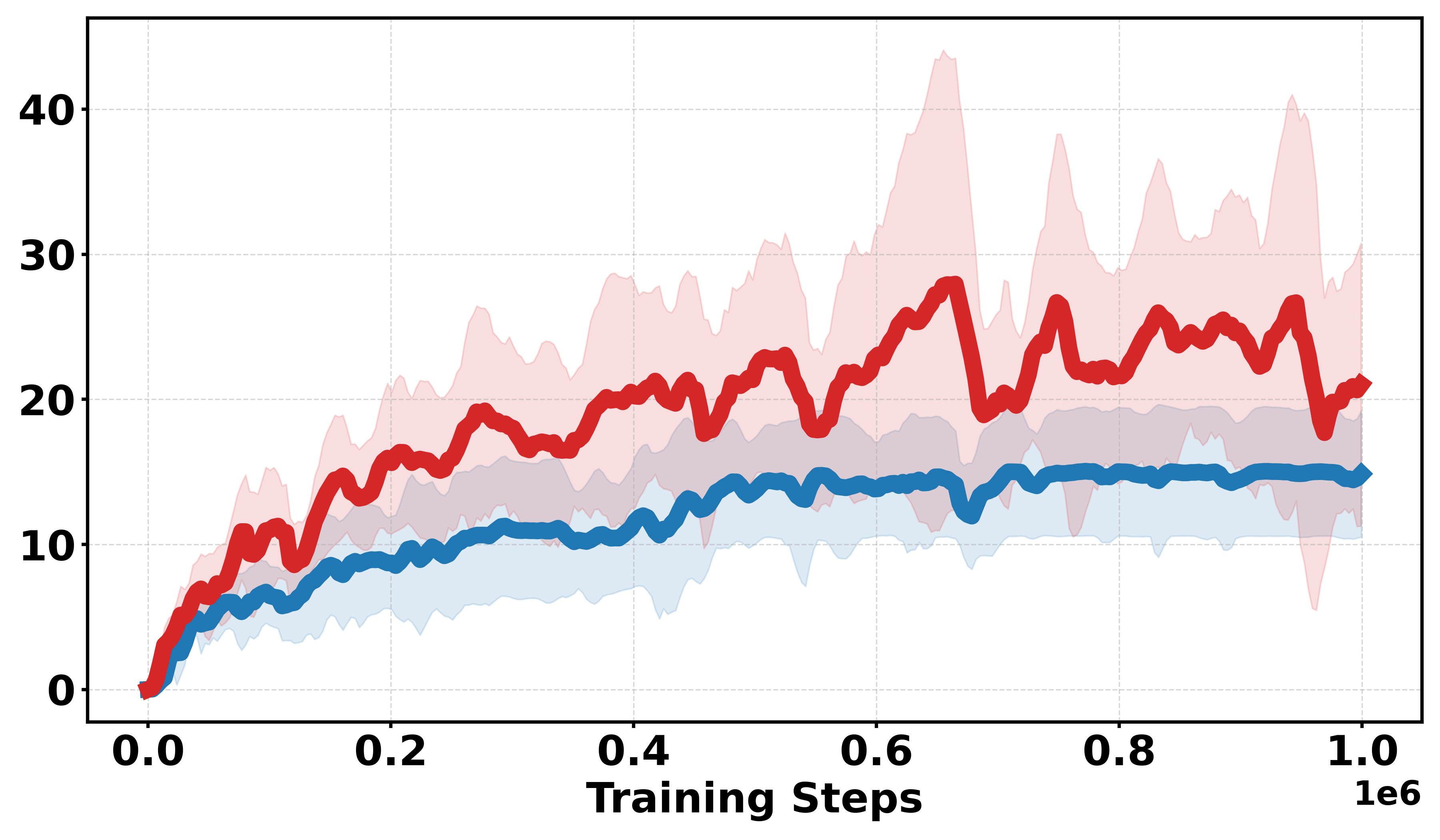}}
\subfloat[Walker2d (Disc.)]{
\includegraphics[width=0.24\textwidth]{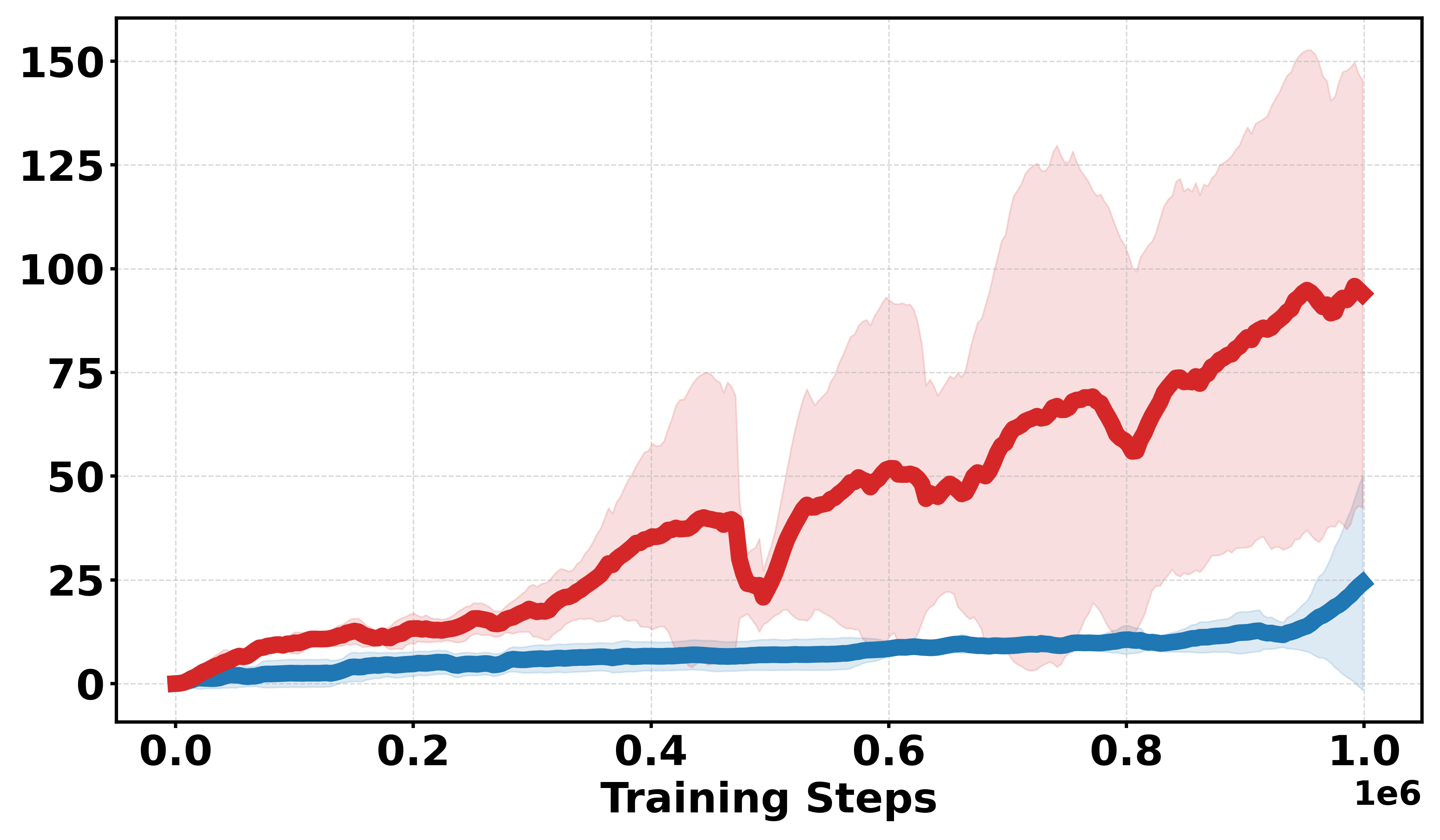}}

\vspace{0.25cm}

\subfloat[Humanoid (Disc.)]{
\includegraphics[width=0.24\textwidth]{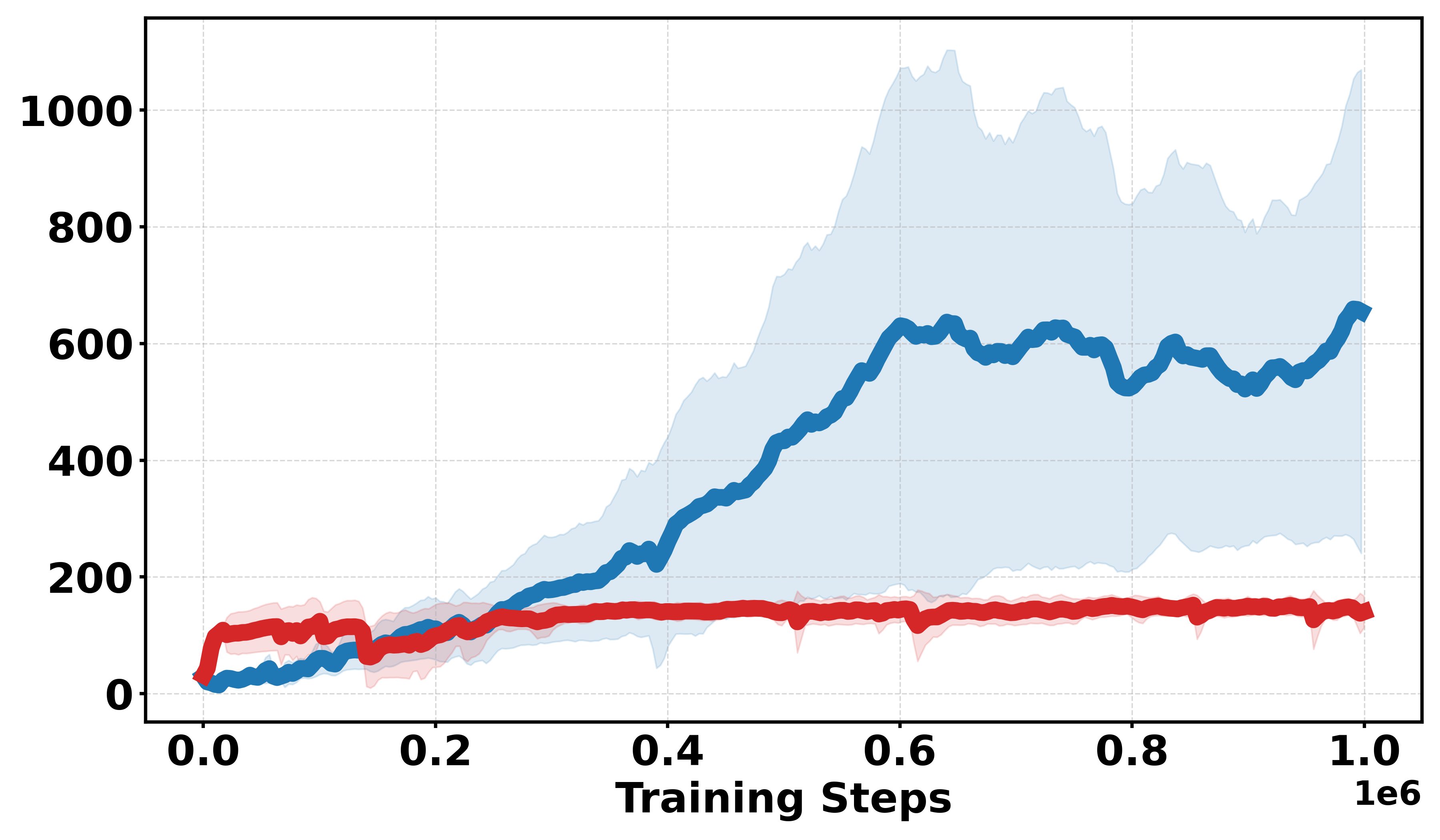}}
\subfloat[Inverted Double Pendulum (Disc.)]{
\includegraphics[width=0.24\textwidth]{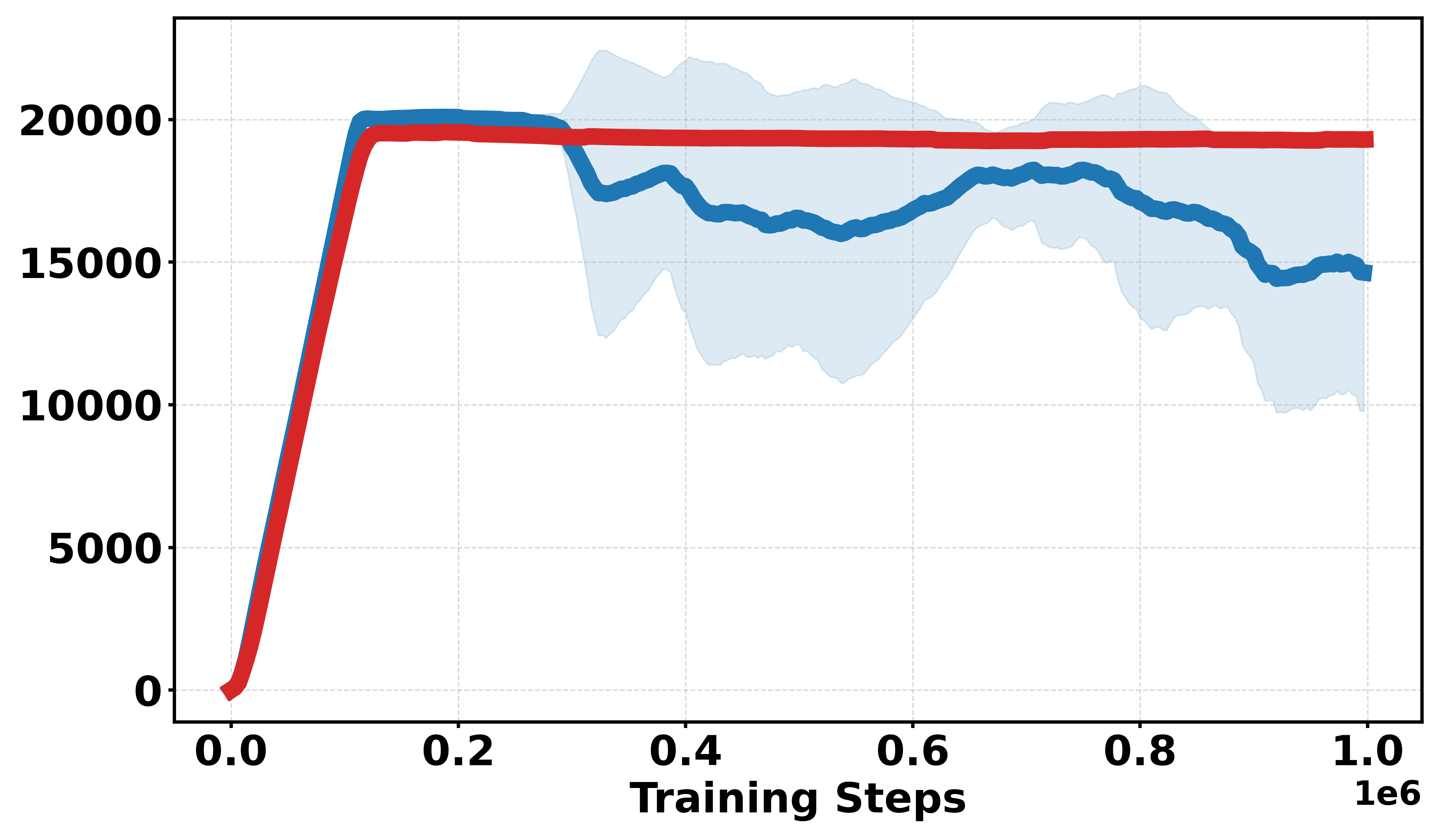}}
\subfloat[Inverted Pendulum (Disc.)]{
\includegraphics[width=0.24\textwidth]{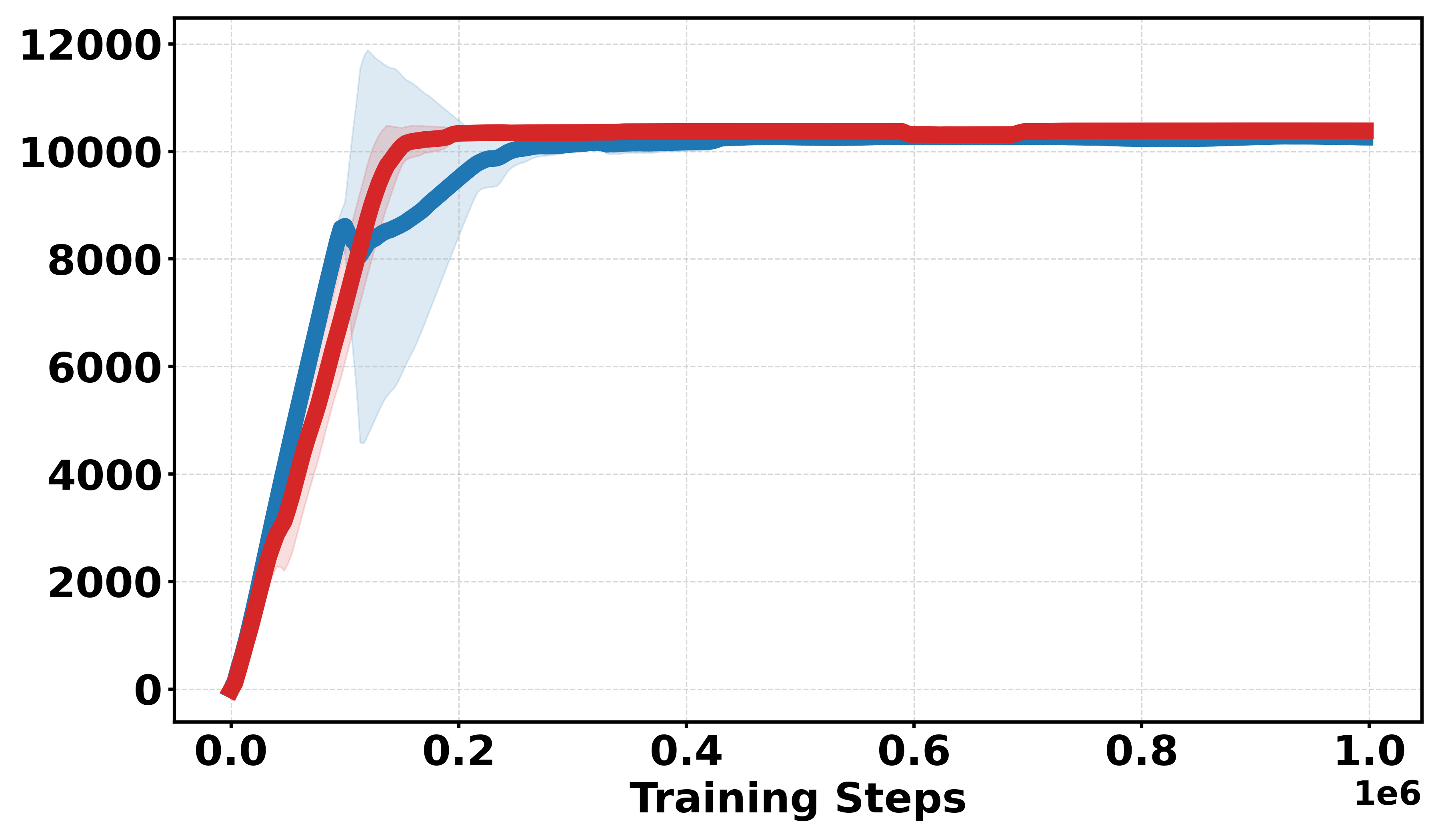}}

\vspace{0cm}

\includegraphics[width=0.6\textwidth]{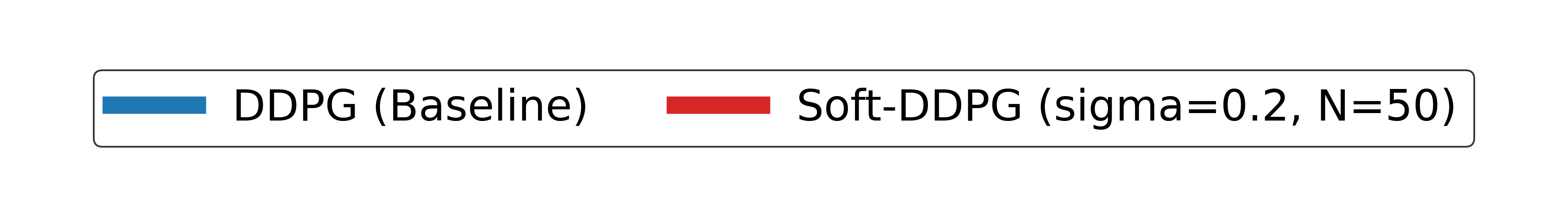}

\vspace{-0.2cm}

\caption{The first two rows show results in the continuous-reward environments,
while the last two rows show the corresponding discrete-reward variants.
Solid lines denote the mean performance over multiple random seeds,
and shaded regions indicate one standard deviation.
}
\label{fig:mujoco_results}

\end{figure*}

\textbf{Continuous reward environments:} \Cref{fig:mujoco_results} (top) and \Cref{tab:combined_results} present the learning curves and final performance averaged over five random seeds. In standard continuous environments where reward signals are dense , vanilla DDPG generally achieves higher performance (e.g., HalfCheetah, Hopper, Walker2d, and Humanoid). This behavior is theoretically expected, as the introduction of GS may introduce approximation bias. Nevertheless, Soft DDPG still achieves competitive results,  outperforming the baseline in specific tasks such as Ant and Inverted Double Pendulum.

\textbf{Discrete reward environments:} The primary strength of Soft DDPG is prominently revealed in environments with non-smooth, discretized reward surfaces (see \Cref{sec:experiment_detail} of the Appendix for the detailed discrete-reward settings). As shown in \Cref{fig:mujoco_results} (bottom) and \Cref{tab:combined_results}, Soft DDPG demonstrates consistent performance gains across most environments, including Ant, HalfCheetah, Hopper, Walker2d, Inverted Pendulum, and Inverted Double Pendulum. In these challenging settings, the standard DDPG suffers from degraded performance due to the non-differentiable nature of the discrete rewards, which induces highly unstable action-gradients. In contrast, operating on the $\sigma$-smoothed Bellman equation allows Soft DDPG to maintain well-defined policy gradients, and it leads to improved stability and higher final returns. 
For additional experimental results, please refer to \Cref{app:discrete_additional} of Appendix.

\begin{table}[h!]
\centering
\footnotesize
\setlength{\tabcolsep}{3pt}
\caption{Performance comparison (mean $\pm$ std over 5 seeds).}
\label{tab:combined_results}
\begin{tabular}{lcccc}
\toprule
\multirow{2}{*}{Environment}
& \multicolumn{2}{c}{Discrete}
& \multicolumn{2}{c}{Continuous} \\
\cmidrule(lr){2-3} \cmidrule(lr){4-5}
& DDPG & Soft DDPG & DDPG & Soft DDPG \\
\midrule
Ant & $190.56 \pm 60.23$ & $\mathbf{271.93 \pm 5.54}$ & $1703.54 \pm 696.72$ & $\mathbf{3742.10 \pm 721.19}$ \\
HalfCheetah & $1599.42 \pm 350.34$ & $\mathbf{1698.09 \pm 223.51}$ & $\mathbf{4275.58 \pm 657.75}$ & $188.81 \pm 1483.14$ \\
Hopper & $14.85 \pm 4.29$ & $\mathbf{22.29 \pm 9.26}$ & $\mathbf{1243.69 \pm 567.00}$ & $883.59 \pm 144.27$ \\
Walker2d & $12.52 \pm 4.13$ & $\mathbf{79.53 \pm 46.71}$ & $\mathbf{2218.03 \pm 1036.51}$ & $1650.24 \pm 381.54$ \\
Humanoid & $\mathbf{580.11 \pm 320.87}$ & $143.89 \pm 14.73$ & $\mathbf{2496.36 \pm 695.18}$ & $437.65 \pm 2.81$ \\
Inverted Pendulum & $10275.08 \pm 53.76$ & $\mathbf{10383.10 \pm 3.35}$ & $\mathbf{892.73 \pm 81.95}$ & $857.69 \pm 132.33$ \\
Inverted Double Pendulum & $14678.94 \pm 4693.22$ & $\mathbf{19284.25 \pm 58.15}$ & $280.40 \pm 34.71$ & $\mathbf{9343.24 \pm 22.42}$ \\
\bottomrule
\end{tabular}
\end{table}

\section{Conclusion and limitations}
\label{sec:conclusion}

In this paper, we proposed Soft-DPG, a principled extension of DPG based on GS at the Bellman operator level. 
By removing the dependence on critic action-gradients, the proposed method ensures well-defined policy updates even for non-smooth Q-functions. 
We further instantiated this framework as Soft DDPG and demonstrated competitive performance on both standard and discretized-reward benchmarks. However, the proposed method introduces a smoothing parameter $\sigma$, which requires careful tuning. 
In addition, as a DDPG-based method, it inherits sensitivity to hyperparameters and lacks convergence guarantees in the deep RL setting.

\bibliographystyle{plainnat}
\bibliography{example_paper}

\newpage
\appendix

\section{Proof}
\subsection{Proof of \Cref{lem:dpg-gaussian}}\label{app:proof-lem-dpg}

We start from the stochastic policy gradient theorem:
\begin{equation}
\nabla_\theta J(\nu_\theta)
=
\frac{1}{1-\gamma}
\sum_{s} d^{\nu_\theta}(s)
\int_a Q^{\nu_\theta}(s,a)\, \nabla_\theta \nu_\theta(a|s)\, da,
\end{equation}
where
\[
d^{\nu_\theta}(s)
=
(1-\gamma)\sum_{t=0}^\infty \gamma^t
P(s_t = s \mid \nu_\theta, s_0 \sim \rho)
\]
is the discounted state visitation distribution under $\nu_\theta$. Consider the Gaussian policy
\begin{equation}
\nu_\theta(a|s)
=
\mathcal{N}(a;\,\pi_\theta(s),\,\sigma^2 I),
\end{equation}
which admits the reparameterization
\begin{equation}
a = \pi_\theta(s) + \sigma w,
\qquad
w \sim \mathcal{N}(0,I).
\end{equation}

The gradient of the policy is given by
\begin{equation}
\nabla_\theta \nu_\theta(a|s)
=
\nu_\theta(a|s)\,
\frac{1}{\sigma^2}
\nabla_\theta \pi_\theta(s)\,(a - \pi_\theta(s)).
\end{equation}

Substituting this into the standard stochastic policy gradient theorem yields:
\begin{align}
\nabla_\theta J(\nu_\theta)
&=
\frac{1}{1-\gamma}
\mathbb{E}_{s \sim d^{\nu_\theta},\, a \sim \nu_\theta}
\left[
Q^{\nu_\theta}(s,a)\,
\frac{1}{\sigma^2}
\nabla_\theta \pi_\theta(s)\,(a - \pi_\theta(s))
\right].
\end{align}

Using the reparameterization $a = \pi_\theta(s) + \sigma w$, we obtain
\begin{align}
\nabla_\theta J(\nu_\theta)
&=
\frac{1}{1-\gamma}
\mathbb{E}_{s \sim d^{\nu_\theta},\, a \sim \nu_\theta}
\left[
\nabla_\theta \pi_\theta(s)\,
\frac{Q^{\nu_\theta}(s,\pi_\theta(s)+\sigma w)\, w}{\sigma}
\right],
\end{align}
which proves the first statement.

\medskip

For the second statement, by \Cref{lem:smoothed_gradient_convergence},
we have the Gaussian smoothing limit
\begin{equation}
\lim_{\sigma \downarrow 0}
\mathbb{E}_{w\sim \mathcal{N}(0,I)}
\left[
\frac{Q^{\nu_\theta}(s,\pi_\theta(s)+\sigma w)\, w}{\sigma}
\right]
=
\nabla_a Q^{\pi_\theta}(s,a)\big|_{a=\pi_\theta(s)}.
\end{equation}

Therefore,
\begin{align}
\lim_{\sigma \downarrow 0}
\nabla_\theta J(\nu_\theta)
&=
\frac{1}{1-\gamma}
\mathbb{E}_{s \sim d^{\nu_\theta}}
\left[
\nabla_\theta \pi_\theta(s)\,
\nabla_a Q^{\pi_\theta}(s,a)\big|_{a=\pi_\theta(s)}
\right],
\end{align}
which recovers the deterministic policy gradient.
\qed

\subsection{Proof of \Cref{lemma:contraction-property}}\label{proof:contraction}
Consider the difference between the two operators:
\begin{align*}
&(T_\sigma^\pi Q)(s,a) - (T_\sigma^\pi Q')(s,a) \\
&\quad= 
\left(
R(s,a)
+ \gamma\, \mathbb{E}_{s',w}
\!\left[ Q\!\left(s', \pi(s') + \sigma w\right) \right]
\right)
-
\left(
R(s,a)
+ \gamma\, \mathbb{E}_{s',w}
\!\left[ Q'\!\left(s', \pi(s') + \sigma w\right) \right]
\right)
\\[0.2cm]
&\quad=
\gamma\, \mathbb{E}_{s' \sim P(\cdot|s,a),\, w \sim \mathcal{N}(0,I)}
\!\left[
Q\!\left(s', \pi(s') + \sigma w\right)
-
Q'\!\left(s', \pi(s') + \sigma w\right)
\right].
\end{align*}

Taking absolute values, we obtain
\begin{align*}
\left|
(T_\sigma^\pi Q)(s,a) - (T_\sigma^\pi Q')(s,a)
\right|
&=
\gamma \left|
\mathbb{E}_{s',w}
\left[
Q\!\left(s', \pi(s') + \sigma w\right)
-
Q'\!\left(s', \pi(s') + \sigma w\right)
\right]
\right|.
\end{align*}

Using Jensen's inequality (or $|\mathbb{E}[X]| \le \mathbb{E}[|X|]$),
\begin{align*}
\left|
(T_\sigma^\pi Q)(s,a) - (T_\sigma^\pi Q')(s,a)
\right|
&\le
\gamma\,
\mathbb{E}_{s',w}
\left[
\left|
Q\!\left(s', \pi(s') + \sigma w\right)
-
Q'\!\left(s', \pi(s') + \sigma w\right)
\right|
\right].
\end{align*}

For every $(s',a')$, the following inequality holds:
\[
\left|
Q(s',a') - Q'(s',a')
\right|
\le
\|Q - Q'\|_\infty.
\]
Therefore, we have
\begin{align*}
\left|
(T_\sigma^\pi Q)(s,a) - (T_\sigma^\pi Q')(s,a)
\right|
&\le
\gamma\,
\mathbb{E}_{s',w}
\left[
\|Q - Q'\|_\infty
\right]
\\
&=
\gamma\, \|Q - Q'\|_\infty.
\end{align*}

Taking the supremum over all $(s,a)$ yields
\begin{align*}
\|T_\sigma^\pi Q - T_\sigma^\pi Q'\|_\infty
&=
\sup_{(s,a)\in\mathcal{S}\times\mathcal{A}}
\left|
(T_\sigma^\pi Q)(s,a) - (T_\sigma^\pi Q')(s,a)
\right|
\\
&\le
\gamma\, \|Q - Q'\|_\infty.
\end{align*}

Hence, $T_\sigma^\pi$ is a $\gamma$-contraction under the supremum norm.

\subsection{Simulation lemma}\label{app:simulation-lemma}
In this subsection, we state a standard simulation lemma for discounted MDPs.
This lemma quantifies how errors in the reward function and transition kernel
propagate to the value function under a fixed policy.
It will be used in the proof of \Cref{lem:smoothing-v-error} to bound the
difference between the original value function $V^\pi$ and the smoothed value
function $V_\sigma^\pi$.
\begin{lemma}\citep{agarwal2019reinforcement}
\label{lem:simulation-lemma}
Let 
$\mathcal{M}=(\mathcal{S},\mathcal{A},P,R,\gamma)$
and
$\widehat{\mathcal{M}}=(\mathcal{S},\mathcal{A},\widehat P,\widehat R,\gamma)$
be two MDPs with the same state space, action space, and discount factor.
Suppose that
\[
\max_{s\in\mathcal{S},\,a\in\mathcal{A}}
|\widehat R(s,a)-R(s,a)|
\le
\epsilon_R
\]
and
\[
\max_{s\in\mathcal{S},\,a\in\mathcal{A}}
\|\widehat P(\cdot\mid s,a)-P(\cdot\mid s,a)\|_1
\le
\epsilon_P .
\]
Then, for any policy $\pi:\mathcal{S}\to\mathcal{A}$,
\[
\|V_{\widehat{\mathcal{M}}}^{\pi}
-
V_{\mathcal{M}}^{\pi}\|_\infty
\le
\frac{\epsilon_R}{1-\gamma}
+
\frac{\gamma\epsilon_P V_{\max}}{2(1-\gamma)},
\]
where
\[
V_{\max}:=\frac{R_{\max}}{1-\gamma}.
\]
\end{lemma}

\subsection{Proof of \Cref{lem:smoothing-v-error}}
\label{proof:bounding_V_error}

Let $m$ denote the dimension of the action space, and let
$w\sim\mathcal{N}(0,I)$. 
Recall that
\[
V_\sigma^\pi(s)
=
\mathbb{E}_{w\sim\mathcal{N}(0,I)}
\left[
Q_\sigma^\pi(s,\pi(s)+\sigma w)
\right].
\]
By expanding $Q_\sigma^\pi$ using the $\sigma$-smoothed Bellman equation, we obtain
\[
V_\sigma^\pi(s)
=
\mathbb{E}_{w\sim\mathcal{N}(0,I)}
\left[
R(s,\pi(s)+\sigma w)
+
\gamma
\int_{\mathcal{S}}
V_\sigma^\pi(s')
P(s'\mid s,\pi(s)+\sigma w)
\,ds'
\right].
\]
Using linearity of expectation, this can be written as
\[
V_\sigma^\pi(s)
=
\mathbb{E}_{w\sim\mathcal{N}(0,I)}
\left[
R(s,\pi(s)+\sigma w)
\right]
+
\gamma
\int_{\mathcal{S}}
V_\sigma^\pi(s')
\mathbb{E}_{w\sim\mathcal{N}(0,I)}
\left[
P(s'\mid s,\pi(s)+\sigma w)
\right]
\,ds' .
\]
Define the smoothed reward function and transition kernel by
\[
R_\sigma(s,a)
:=
\mathbb{E}_{w\sim\mathcal{N}(0,I)}
\left[
R(s,a+\sigma w)
\right],
\]
and
\[
P_\sigma(s'\mid s,a)
:=
\mathbb{E}_{w\sim\mathcal{N}(0,I)}
\left[
P(s'\mid s,a+\sigma w)
\right].
\]
Then
\[
V_\sigma^\pi(s)
=
R_\sigma(s,\pi(s))
+
\gamma
\int_{\mathcal{S}}
V_\sigma^\pi(s')
P_\sigma(s'\mid s,\pi(s))
\,ds' .
\]
Hence, $V_\sigma^\pi$ is the value function of the modified MDP
$\mathcal{M}_\sigma
=
(\mathcal{S},\mathcal{A},P_\sigma,R_\sigma,\gamma)$.

Applying the simulation lemma in \Cref{lem:simulation-lemma} with
$\widehat{\mathcal M}=\mathcal M_\sigma$,
$\widehat R=R_\sigma$, and $\widehat P=P_\sigma$, we obtain
\begin{equation}
\label{eqn:simulation-lemma}
\|V^\pi - V_\sigma^\pi\|_\infty
\le
\frac{1}{1-\gamma}
\|R-R_\sigma\|_\infty
+
\frac{\gamma V_{\max}}{2(1-\gamma)}
\|P-P_\sigma\|_{\infty,1}.
\end{equation}
Here,
\[
\|R-R_\sigma\|_\infty
:=
\sup_{(s,a)\in\mathcal{S}\times\mathcal{A}}
|R(s,a)-R_\sigma(s,a)|,
\]
and
\[
\|P-P_\sigma\|_{\infty,1}
:=
\sup_{(s,a)\in\mathcal{S}\times\mathcal{A}}
\int_{\mathcal{S}}
\left|
P(s'\mid s,a)-P_\sigma(s'\mid s,a)
\right|
\,ds' .
\]

We now bound the two error terms. 
Assume that the reward function is $L_R$-Lipschitz continuous with respect to
the Euclidean norm on the action space; that is, for all
$s\in\mathcal{S}$ and $a,a'\in\mathcal{A}$,
\begin{equation}
\label{eqn:reward-lipschitz}
|R(s,a)-R(s,a')|
\le
L_R\|a-a'\|_2 .
\end{equation}
Then
\begin{align}
\|R-R_\sigma\|_\infty
&=
\sup_{(s,a)\in\mathcal{S}\times\mathcal{A}}
\left|
R(s,a)
-
\mathbb{E}_{w\sim\mathcal{N}(0,I)}
\left[
R(s,a+\sigma w)
\right]
\right|
\nonumber\\
&\le
\sup_{(s,a)\in\mathcal{S}\times\mathcal{A}}
\mathbb{E}_{w\sim\mathcal{N}(0,I)}
\left[
|R(s,a)-R(s,a+\sigma w)|
\right]
\nonumber\\
&\le
\mathbb{E}_{w\sim\mathcal{N}(0,I)}
\left[
L_R\sigma\|w\|_2
\right]
\nonumber\\
&=
L_R\sigma
\mathbb{E}_{w\sim\mathcal{N}(0,I)}
\left[
\|w\|_2
\right]
\nonumber\\
&\le
L_R\sigma \sqrt{m}.
\label{eqn:reward-error-bound}
\end{align}
The last inequality follows from Jensen's inequality and the properties of the standard multivariate normal distribution \citep{bishop2006pattern}:
\[
\mathbb{E}_{w\sim\mathcal{N}(0,I)}[\|w\|_2]
\le
\sqrt{
\mathbb{E}_{w\sim\mathcal{N}(0,I)}[\|w\|_2^2]
}
=
\sqrt{m},
\]

Next, assume that the transition kernel is $L_P$-Lipschitz continuous with
respect to the action argument in the $L_1$ sense; that is, for all
$s\in\mathcal{S}$ and $a,a'\in\mathcal{A}$,
\begin{equation}
\label{eqn:transition-lipschitz}
\int_{\mathcal{S}}
\left|
P(s'\mid s,a)-P(s'\mid s,a')
\right|
\,ds'
\le
L_P\|a-a'\|_2 .
\end{equation}
Using this assumption,
\begin{align}
\|P-P_\sigma\|_{\infty,1}
&=
\sup_{(s,a)\in\mathcal{S}\times\mathcal{A}}
\int_{\mathcal{S}}
\left|
P(s'\mid s,a)
-
\mathbb{E}_{w\sim\mathcal{N}(0,I)}
\left[
P(s'\mid s,a+\sigma w)
\right]
\right|
\,ds'
\nonumber\\
&\le
\sup_{(s,a)\in\mathcal{S}\times\mathcal{A}}
\mathbb{E}_{w\sim\mathcal{N}(0,I)}
\left[
\int_{\mathcal{S}}
\left|
P(s'\mid s,a)-P(s'\mid s,a+\sigma w)
\right|
\,ds'
\right]
\nonumber\\
&\le
\mathbb{E}_{w\sim\mathcal{N}(0,I)}
\left[
L_P\sigma\|w\|_2
\right]
\nonumber\\
&=
L_P\sigma \sqrt{m}.
\label{eqn:transition-error-bound}
\end{align}

Substituting \eqref{eqn:reward-error-bound} and
\eqref{eqn:transition-error-bound} into
\eqref{eqn:simulation-lemma}, we obtain
\[
\|V^\pi - V_\sigma^\pi\|_\infty
\le
\frac{L_R\sigma \sqrt{m}}{1-\gamma}
+
\frac{\gamma V_{\max}}{2(1-\gamma)}
L_P\sigma \sqrt{m} .
\]
Therefore,
\[
\|V^\pi - V_\sigma^\pi\|_\infty
\le
\frac{\sigma \sqrt{m}}{1-\gamma}
\left(
L_R
+
\frac{\gamma}{2}L_PV_{\max}
\right).
\]
This completes the proof.

\subsection{Proof of \Cref{lem:q-lipschitz}}\label{proof:q-lipschitz}

Recall the definition of the smoothed action-value function:
\[
Q_\sigma^\pi(s,a) = R(s,a) + \gamma \int_{\mathcal{S}} V_\sigma^\pi(s') P(s' \mid s,a) ds'.
\]
For any state $s \in \mathcal{S}$ and two actions $a, a' \in \mathcal{A}$, we can bound the difference as follows:
\begin{align*}
\left| Q_\sigma^\pi(s,a) - Q_\sigma^\pi(s,a') \right| 
&\le 
\left| R(s,a) - R(s,a') \right| \\
&\quad + \gamma \int_{\mathcal{S}} V_\sigma^\pi(s') \left| P(s' \mid s,a) - P(s' \mid s,a') \right| ds'.
\end{align*}

By the assumptions in \Cref{lem:smoothing-v-error}, the reward function is $L_R$-Lipschitz continuous, which gives:
\begin{equation}\label{eq:q_lip_reward}
\left| R(s,a) - R(s,a') \right| \le L_R \|a - a'\|_2.
\end{equation}

For the second term, we know that the value function is bounded by $V_{\max}$, i.e., $V_\sigma^\pi(s') \le V_{\max} = R_{\max}/(1-\gamma)$. Furthermore, since the transition density is assumed to be $L_P$-Lipschitz continuous with respect to the action in the $L_1$ sense, we have:
\begin{align}\label{eq:q_lip_transition}
\int_{\mathcal{S}} V_\sigma^\pi(s') \left| P(s' \mid s,a) - P(s' \mid s,a') \right| ds' 
&\le 
V_{\max} \int_{\mathcal{S}} \left| P(s' \mid s,a) - P(s' \mid s,a') \right| ds' \nonumber \\
&\le 
V_{\max} L_P \|a - a'\|_2.
\end{align}

Combining \eqref{eq:q_lip_reward} and \eqref{eq:q_lip_transition} yields:
\begin{align*}
\left| Q_\sigma^\pi(s,a) - Q_\sigma^\pi(s,a') \right| 
&\le 
L_R \|a - a'\|_2 + \gamma V_{\max} L_P \|a - a'\|_2 \\
&= 
\left( L_R + \gamma L_P V_{\max} \right) \|a - a'\|_2.
\end{align*}

This demonstrates that for any given state $s$, the smoothed action-value function $Q_\sigma^\pi(s,\cdot)$ is $L_Q$-Lipschitz continuous with respect to the action, where the Lipschitz constant is given by $L_Q = L_R + \gamma L_P V_{\max}$. This completes the proof.

\subsection{Proof of \Cref{lem:smoothing-error}}\label{proof:bounding_Q_error}
Let $m$ denote the dimension of the action space. Since $Q^\pi = T^\pi Q^\pi$ and $Q_\sigma^\pi = T_\sigma^\pi Q_\sigma^\pi$, 
\[
\|Q^\pi - Q_\sigma^\pi\|_\infty
= \|T^\pi Q^\pi - T_\sigma^\pi Q_\sigma^\pi\|_\infty
\le 
\|T^\pi Q^\pi - T^\pi Q_\sigma^\pi\|_\infty
+
\|T^\pi Q_\sigma^\pi - T_\sigma^\pi Q_\sigma^\pi\|_\infty.
\]
The first term is bounded by the contraction of $T^\pi$:
\begin{align}\label{eq:contraction1}
\|T^\pi Q^\pi - T^\pi Q_\sigma^\pi\|_\infty
\le \gamma \|Q^\pi - Q_\sigma^\pi\|_\infty.
\end{align}

For the second term, note that the reward and transition kernel are not smoothed; 
the only difference is in the value of the next action:
\[
(T^\pi Q)(s,a)
= R(s,a) + \gamma \mathbb{E}_{s'\sim P(\cdot|s,a)}
\left[ Q(s',\pi(s')) \right],
\]
\[
(T_\sigma^\pi Q)(s,a)
= R(s,a) + \gamma 
\mathbb{E}_{s'\sim P(\cdot|s,a),w\sim\mathcal{N}(0,I)}
\left[ Q(s',\pi(s')+\sigma w) \right].
\]
By assumption, for every $s\in\mathcal{S}$ and $a,a'\in\mathcal{A}$,
\[
\left|
Q_\sigma^\pi(s,a)-Q_\sigma^\pi(s,a')
\right|
\le
L_Q\|a-a'\|_2 .
\]
Thus, taking $a=\pi(s')$ and $a'=\pi(s')+\sigma w$, we obtain
\[
\left|
Q_\sigma^\pi(s',\pi(s'))
-
Q_\sigma^\pi(s',\pi(s')+\sigma w)
\right|
\le
L_Q\|\sigma w\|_2
=
L_Q\sigma\|w\|_2 .
\]
Therefore, for any $(s,a)\in\mathcal{S}\times\mathcal{A}$,
\begin{align}
&\left|
(T^\pi Q_\sigma^\pi)(s,a)
-
(T_\sigma^\pi Q_\sigma^\pi)(s,a)
\right| \nonumber\\
&=
\gamma
\left|
\mathbb{E}_{s'\sim P(\cdot|s,a)}
\mathbb{E}_{w\sim\mathcal{N}(0,I)}
\left[
Q_\sigma^\pi(s',\pi(s'))
-
Q_\sigma^\pi(s',\pi(s')+\sigma w)
\right]
\right| \nonumber\\
&\le
\gamma
\mathbb{E}_{s'\sim P(\cdot|s,a)}
\mathbb{E}_{w\sim\mathcal{N}(0,I)}
\left[
\left|
Q_\sigma^\pi(s',\pi(s'))
-
Q_\sigma^\pi(s',\pi(s')+\sigma w)
\right|
\right] \nonumber\\
&\le
\gamma
\mathbb{E}_{s'\sim P(\cdot|s,a)}
\mathbb{E}_{w\sim\mathcal{N}(0,I)}
\left[
L_Q\sigma\|w\|_2
\right] \nonumber\\
&\le
\gamma
\mathbb{E}_{s'\sim P(\cdot|s,a)}
\mathbb{E}_{w\sim\mathcal{N}(0,I)}
\left[
L_Q\sigma\|w\|_2
\right] \nonumber\\
&=
\gamma L_Q\sigma
\mathbb{E}_{w\sim\mathcal{N}(0,I)}[\|w\|_2] \nonumber\\
&\le
\gamma L_Q\sigma \sqrt{m}.
\end{align}
The last inequality follows from Jensen's inequality and the properties of the standard multivariate normal distribution \citep{bishop2006pattern}:
$
\mathbb{E}_{w\sim\mathcal{N}(0,I)}[\|w\|_2]
\le
\sqrt{
\mathbb{E}_{w\sim\mathcal{N}(0,I)}[\|w\|_2^2]
}
=
\sqrt{m}.
$
Since this bound holds for every $(s,a)$, taking the supremum over $(s,a)$ yields
\begin{align}\label{eq:contraction2}
\|T^\pi Q_\sigma^\pi - T_\sigma^\pi Q_\sigma^\pi\|_\infty
\le
\gamma L_Q\sigma \sqrt{m} .
\end{align}

Combining \eqref{eq:contraction1} and \eqref{eq:contraction2}, we obtain
\[
\|Q^\pi - Q_\sigma^\pi\|_\infty
\le
\gamma\|Q^\pi - Q_\sigma^\pi\|_\infty
+
\gamma L_Q\sigma \sqrt{m}.
\]
Rearranging gives
\[
\|Q^\pi - Q_\sigma^\pi\|_\infty
\le
\frac{\gamma L_Q\sigma \sqrt{m}}{1-\gamma}.
\]
This completes the proof.

\subsection{Proof of \Cref{lemma:gradient-Bellman-eq}}\label{proof:gradient_flow}
Using the chain rule, we have
\begin{align*}
{\nabla _\theta }V_\sigma ^{{\pi _\theta }}(s) =& {\nabla _\theta }{E_{w \sim N(0,I)}}[Q_\sigma ^{{\pi _\theta }}(s,{\pi _\theta }(s) + \sigma w)]\\
=& {\left. {{\nabla _\theta }{E_{w \sim N(0,I)}}[Q_\sigma ^{{\pi _\theta }}(s,a + \sigma w)]} \right|_{a = {\pi _\theta }(s)}} + {\left. {{\nabla _\theta }{E_{w \sim N(0,I)}}[Q_\sigma ^\pi (s,{\pi _\theta }(s) + \sigma w)]} \right|_{\pi  = {\pi _\theta }}},\\
\end{align*}
where
\begin{align*}
{\left. {{\nabla _\theta }{E_{w \sim N(0,I)}}[Q_\sigma ^{{\pi _\theta }}(s,a + \sigma w)]} \right|_{a = {\pi _\theta }(s)}} =& {\left. {{\nabla _\theta }\left\{ {{R_\sigma }(s,a) + \gamma \int_{s' \in S} {V_\sigma ^{{\pi _\theta }}(s'){P_\sigma }(s'|s,a)ds'} } \right\}} \right|_{a = {\pi _\theta }(s)}}\\
=& \gamma \int_{s' \in S} {{\nabla _\theta }V_\sigma ^{{\pi _\theta }}(s'){P_\sigma }(s'|s,{\pi _\theta }(s))ds'}.
\end{align*}

This completes the proof.

\subsection{Proof of \Cref{thm:smoothed-dpg-thm}}\label{proof:SDPG}
The objective function is defined as the expected return from the initial state distribution $\rho_0$:$$J_{\sigma}^{\pi_\theta} = \mathbb{E}_{s_0 \sim \rho_0} \big[ V_{\sigma}^{\pi_\theta}(s_0) \big].$$Taking the gradient with respect to $\theta$ yields:$$\nabla_\theta J_{\sigma}^{\pi_\theta} = \int_{\mathcal{S}} \rho_0(s_0) \nabla_\theta V_{\sigma}^{\pi_\theta}(s_0) ds_0.$$

By substituting the gradient Bellman equation (\Cref{lemma:gradient-Bellman-eq}) into the above expression, we obtain the first unrolling step:
\begin{align*}
\nabla_\theta J_{\sigma}^{\pi_\theta}&= \int_{\mathcal{S}} \rho_0(s_0) \Bigl(\left. \nabla_\theta \mathbb{E}_{w \sim \mathcal{N}(0,I)} \big[ Q_{\sigma}^{\pi_\theta}(s_0,\pi_\theta(s_0)+\sigma w) \big] \right|_{\pi=\pi\theta}\\&\quad+\gamma \int_{\mathcal{S}} P_{\sigma}(s_1|s_0,\pi_\theta(s_0)) \nabla_\theta V_{\sigma}^{\pi_\theta}(s_1) ds_1\Bigr) ds_0 \\&= \int_{\mathcal{S}} \rho_0(s) \left. \nabla_\theta \mathbb{E}_{w \sim \mathcal{N}(0,I)} \big[ Q_{\sigma}^{\pi_\theta}(s,\pi_\theta(s)+\sigma w) \big] \right|_{\pi=\pi\theta} ds_0+\gamma \int_{\mathcal{S}} \rho_1(s_1) \nabla_\theta V_{\sigma}^{\pi_\theta}(s_1) ds_1,
\end{align*}
where $\rho_1(s_1) = \int_{\mathcal{S}} \rho_0(s_0) P_{\sigma}(s_1|s_0,\pi_\theta(s_0)) ds_0$ is the state distribution at time step 1.We can recursively substitute the gradient Bellman equation for $\nabla_\theta V_{\sigma}^{\pi_\theta}(s_1)$ to reveal the second unrolling step:$$\begin{aligned}\nabla_\theta J_{\sigma}^{\pi_\theta}&= \int_{\mathcal{S}} \rho_0(s) \left. \nabla_\theta \mathbb{E}_{w \sim \mathcal{N}(0,I)} \big[ Q_{\sigma}^{\pi_\theta}(s,\pi_\theta(s)+\sigma w) \big] \right|_{\pi=\pi\theta} ds \\&\quad + \gamma \int_{\mathcal{S}} \rho_1(s_1) \Bigl( \left. \nabla_\theta \mathbb{E}{w \sim \mathcal{N}(0,I)} \big[ Q_{\sigma}^{\pi_\theta}(s_1,\pi_\theta(s_1)+\sigma w) \big] \right|_{\pi=\pi\theta} \\&\quad+ \gamma \int_{\mathcal{S}} P_{\sigma}(s_2|s_1,\pi_\theta(s_1)) \nabla_\theta V_{\sigma}^{\pi_\theta}(s_2) ds_2 \Bigr) ds_1 \\&= \int_{\mathcal{S}} \rho_0(s) \left. \nabla_\theta \mathbb{E}_{w \sim \mathcal{N}(0,I)} \big[ Q_{\sigma}^{\pi_\theta}(s,\pi_\theta(s)+\sigma w) \big] \right|_{\pi=\pi\theta} ds \\&\quad + \gamma \int_{\mathcal{S}} \rho_1(s) \left. \nabla_\theta \mathbb{E}_{w \sim \mathcal{N}(0,I)} \big[ Q_{\sigma}^{\pi_\theta}(s,\pi_\theta(s)+\sigma w) \big] \right|_{\pi=\pi\theta} ds+\gamma^2 \int_{\mathcal{S}} \rho_2(s) \nabla_\theta V_{\sigma}^{\pi_\theta}(s) ds\end{aligned}$$where $\rho_2(s) = \int_{\mathcal{S}} \rho_1(s_1) P_{\sigma}(s|s_1,\pi_\theta(s_1)) ds_1$.

By continuing this recursive unrolling process for infinite time steps, the terms form a geometric series weighted by the state distributions at each step $t$:$$\nabla_\theta J_{\sigma}^{\pi_\theta} = \int_{\mathcal{S}} \sum_{t=0}^{\infty} \gamma^t \rho_t(s) \left. \nabla_\theta \mathbb{E}_{w \sim \mathcal{N}(0,I)} \big[ Q_{\sigma}^{\pi_\theta}(s,\pi_\theta(s)+\sigma w) \big] \right|_{\pi=\pi_\theta} ds.$$Recognizing that the discounted state visitation distribution is defined as $\rho^{\nu_\theta}(s) = \sum_{t=0}^{\infty} \gamma^t \rho_t(s)$, this simplifies compactly to an expectation over $\rho^{\nu_\theta}$:$$\nabla_\theta J_{\sigma}^{\pi_\theta} = \left. \mathbb{E}_{s \sim \rho^{\nu_\theta},\, w \sim \mathcal{N}(0,I)} \big[ \nabla_\theta Q_{\sigma}^{\pi_\theta}(s, \pi_\theta(s)+\sigma w) \big] \right|_{\pi=\pi_\theta}.$$From here, we proceed with the derivation of the policy gradient using the chain rule and the Gaussian log-derivative trick:

Moreover, we have
\begin{align*}
\nabla_\theta J_{\sigma}^{\pi_\theta}
=&\;
\left.
\mathbb{E}_{w \sim \mathcal{N}(0,I),\, s \sim \rho^{\nu_\theta}}
\big[
\nabla_\theta 
Q_{\sigma}^{\pi_\theta}(s, \pi_\theta(s)+\sigma w)
\big]
\right|_{\pi=\pi_\theta}
\\[4pt]
=&\;
\mathbb{E}_{w \sim \mathcal{N}(0,I),\, s \sim \rho^{\nu_\theta}}
\left[
\nabla_\theta \pi_\theta(s)\,
\left.
\nabla_a 
Q_{\sigma}^{\pi_\theta}(s, a + \sigma w)
\right|_{a=\pi_\theta(s)}
\right]
\\[4pt]
=&\;
\mathbb{E}_{s \sim \rho^{\nu_\theta}}
\left[
\nabla_\theta \pi_\theta(s)
\left.
\nabla_a 
\mathbb{E}_{w \sim \mathcal{N}(0,I)}
\big[
Q_{\sigma}^{\pi_\theta}(s, a + \sigma w)
\big]
\right|_{a=\pi_\theta(s)}
\right]
\\[4pt]
=&\;
\mathbb{E}_{w \sim \mathcal{N}(0,I),\, s \sim \rho^{\nu_\theta}}
\left[
\nabla_\theta \pi_\theta(s)\,
w\,
\frac{
Q_{\sigma}^{\pi_\theta}(s, \pi_\theta(s)+\sigma w)
}{
\sigma
}
\right]
\\[4pt]
=&\;
\mathbb{E}_{w \sim \mathcal{N}(0,I),\, s \sim \rho^{\nu_\theta}}
\left[
\nabla_\theta \pi_\theta(s)\,
\frac{\tilde{a} - \pi_\theta(s)}{\sigma}
\frac{
Q_{\sigma}^{\pi_\theta}(s,\tilde{a})
}{
\sigma
}
\right]
\\[4pt]
=&\;
\mathbb{E}_{s \sim \rho^{\nu_\theta},\, \tilde{a} \sim \nu_\theta}
\left[
-\frac{1}{2\sigma^2}
\nabla_\theta 
\left\| \tilde{a} - \pi_\theta(s) \right\|_2^2\;
Q_{\sigma}^{\pi_\theta}(s,\tilde{a})
\right],
\end{align*}
where the fourth equality follows from the definition of the gradient of Gaussian smoothing. The fifth equality is obtained by applying the change of variable $\tilde{a} = \pi_\theta(s) + \sigma w$. Finally, the last equality uses the gradient identity $\nabla_\theta \|\tilde{a} - \pi_\theta(s)\|_2^2 = -2(\tilde{a} - \pi_\theta(s)) \nabla_\theta \pi_\theta(s)$, with $\nu_\theta(\cdot|s) = \mathcal{N}(\pi_\theta(s), \sigma^2 I)$ denoting the Gaussian smoothing distribution.
This completes the proof.

\subsection{Alternative proof of \Cref{thm:smoothed-dpg-thm} via standard policy gradient}
\label{proof:alt-sdpg}

The result in \Cref{thm:smoothed-dpg-thm} can also be derived more directly
from the standard stochastic policy gradient theorem.
Define the Gaussian policy
\[
\nu_\theta(\tilde a \mid s)
=
\mathcal{N}(\tilde a;\pi_\theta(s),\sigma^2 I).
\]
The $\sigma$-smoothed Bellman equation can be interpreted as the Bellman
equation induced by this Gaussian policy. In particular,
\[
Q_\sigma^{\pi_\theta}(s,a)
=
R(s,a)
+
\gamma
\mathbb{E}_{s'\sim P(\cdot|s,a), a'\sim \nu_\theta(\cdot|s')}
\left[
Q_\sigma^{\pi_\theta}(s',\tilde a')
\right].
\]
Thus, $Q_\sigma^{\pi_\theta}$ coincides with the action-value function
associated with the Gaussian policy $\nu_\theta$.

Let
$
\rho^{\nu_\theta}(s)
:=
\sum_{t=0}^{\infty}\gamma^t \rho_t(s)
$
denote the unnormalized discounted state visitation measure induced by
$\nu_\theta$, where $\rho_t$ is the state distribution at time step $t$.
Then, using the standard stochastic policy gradient theorem with this
unnormalized measure, we obtain
\[
\begin{aligned}
\nabla_\theta J_\sigma
&=
\int_{\mathcal S}
\rho^{\nu_\theta}(s)
\int_{\mathbb R^m}
\nabla_\theta \nu_\theta(\tilde a\mid s)
Q_\sigma^{\pi_\theta}(s,\tilde a)
\,d\tilde a\,ds
\\
&=
\mathbb{E}_{s\sim \rho^{\nu_\theta},\,\tilde a\sim\nu_\theta}
\left[
\nabla_\theta \log \nu_\theta(\tilde a\mid s)
Q_\sigma^{\pi_\theta}(s,\tilde a)
\right].
\end{aligned}
\]

For the Gaussian policy $\nu_\theta(\tilde a\mid s)$, the log-probability is
\[
\log \nu_\theta(\tilde a\mid s)
=
-\frac{1}{2\sigma^2}
\bigl\|\tilde a-\pi_\theta(s)\bigr\|_2^2
-
\frac{m}{2}\log(2\pi\sigma^2).
\]
Taking the gradient with respect to $\theta$ yields
\[
\begin{aligned}
\nabla_\theta \log \nu_\theta(\tilde a\mid s)
=
-\frac{1}{2\sigma^2}
\nabla_\theta
\bigl\|\tilde a-\pi_\theta(s)\bigr\|_2^2
=
\frac{1}{\sigma^2}
\nabla_\theta \pi_\theta(s)
(\tilde a-\pi_\theta(s)).
\end{aligned}
\]
Substituting this identity into the policy gradient expression gives
\[
\begin{aligned}
\nabla_\theta J_\sigma
&=
\mathbb{E}_{s\sim \rho^{\nu_\theta},\,\tilde a\sim\nu_\theta}
\left[
-\frac{1}{2\sigma^2}
\nabla_\theta
\bigl\|\tilde a-\pi_\theta(s)\bigr\|_2^2
Q_\sigma^{\pi_\theta}(s,\tilde a)
\right]
\\
&=
\mathbb{E}_{s\sim \rho^{\nu_\theta},\,\tilde a\sim\nu_\theta}
\left[
\frac{1}{\sigma^2}
\nabla_\theta \pi_\theta(s)
(\tilde a-\pi_\theta(s))
Q_\sigma^{\pi_\theta}(s,\tilde a)
\right].
\end{aligned}
\]
This recovers the Soft-DPG formulation and completes the alternative proof.

\subsection{Sensitivity analysis of smoothing parameters}
\label{sec:sensitivity}

The proposed Soft DDPG introduces two key hyperparameters:
the smoothing scale $\sigma$ and the number of Monte Carlo samples $N$ used to approximate the Gaussian smoothing expectation.
To better understand the influence of these parameters, we conduct a sensitivity analysis.

We perform this analysis on the Ant environment from OpenAI Gym \citep{brockman2016openai} as a representative benchmark.
While the optimal hyperparameters may vary across environments due to differences in dynamics and reward structures,
our objective is not to perform exhaustive per-environment tuning, but to identify a robust configuration that generalizes well across tasks.

\begin{figure}[htbp]
    \centering
    \begin{subfigure}[b]{0.48\textwidth}
        \centering
        \includegraphics[width=\textwidth]{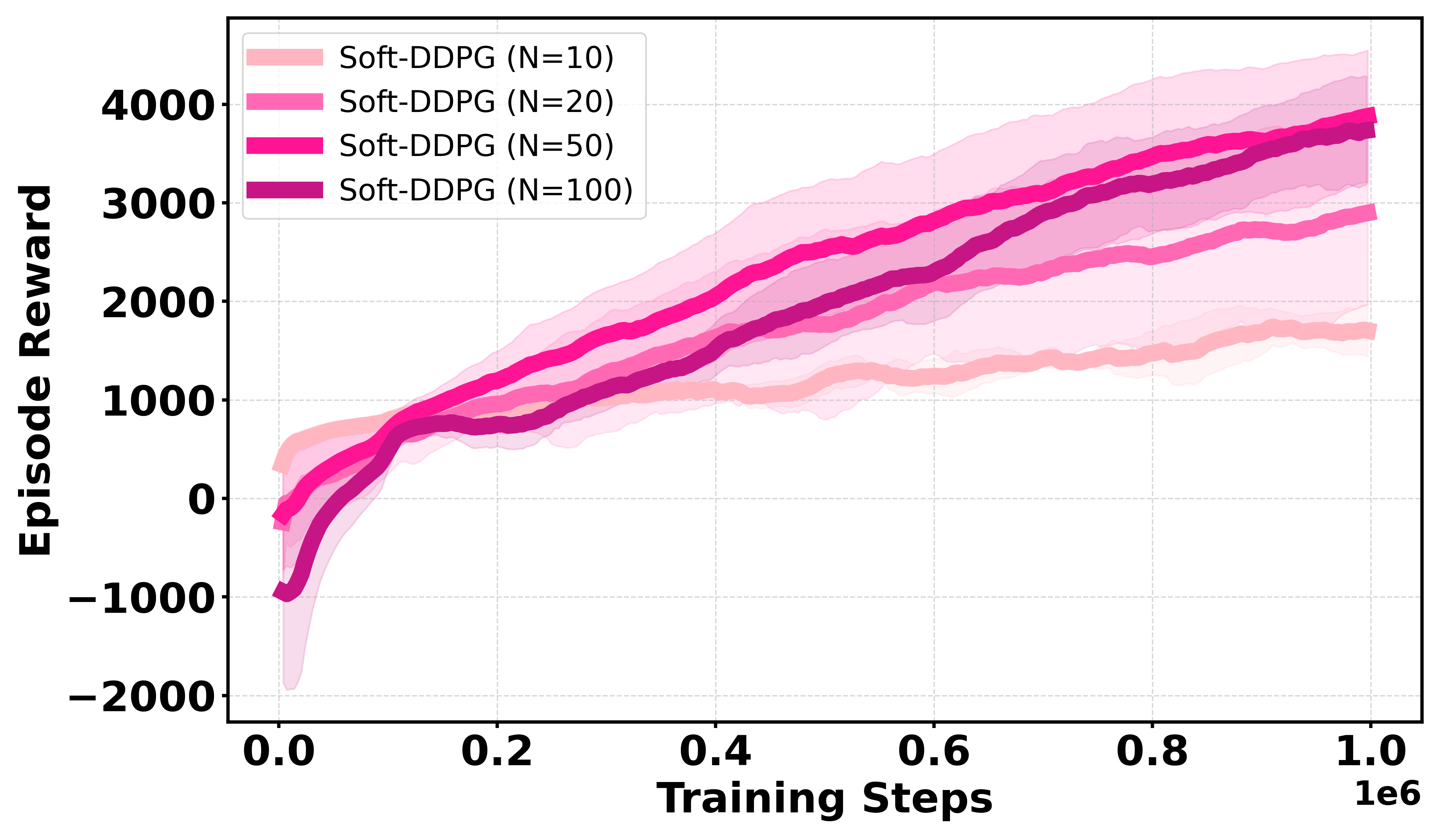}
        \caption{Sensitivity to the number of samples $N$ ($\sigma=0.1$)}
        \label{fig:ant_ablation_n}
    \end{subfigure}
    \hfill 
    \begin{subfigure}[b]{0.48\textwidth}
        \centering
        \includegraphics[width=\textwidth]{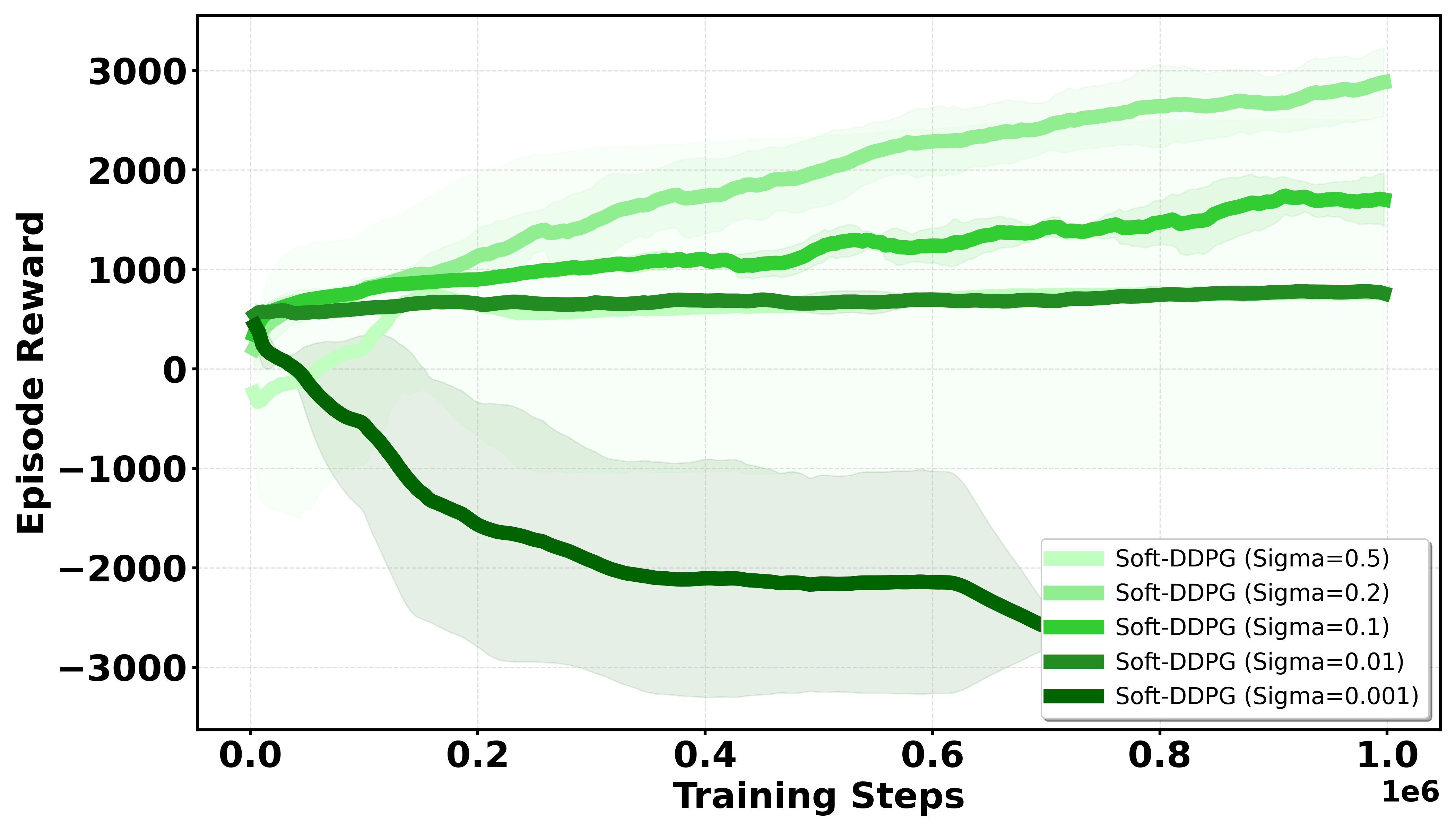}
        \caption{Sensitivity to the smoothing scale $\sigma$ ($N=10$)}
        \label{fig:ant_ablation_sigma}
    \end{subfigure}
    
    \caption{
Sensitivity analysis of Soft DDPG hyperparameters on the Ant environment.
(Left) Performance as a function of the number of samples $N$ with $\sigma=0.2$.
(Right) Performance as a function of the smoothing scale $\sigma$ with $N=10$.
The results show that $N=50$ and $\sigma=0.2$ provide the best trade-off between performance and stability.
}
    \label{fig:ant_sensitivity_analysis}
\end{figure}

\paragraph{Effect of the number of samples $N$.}
We first evaluate the impact of the Monte Carlo sample size by fixing the smoothing scale to $\sigma = 0.2$ and varying $N \in \{10, 20, 50, 100\}$.
As shown in \Cref{fig:ant_ablation_n}, increasing $N$ initially improves performance, as the approximation of the smoothed Bellman expectation becomes more accurate.
However, beyond a certain point, the gains saturate and even slightly degrade due to increased variance and computational overhead.
Empirically, we observe that $N=50$ achieves the best overall performance, providing a favorable trade-off between approximation accuracy and efficiency.
While the optimal choice of $N$ may vary across environments, we find that $N=50$ provides a robust and effective setting in practice.
Therefore, we use $N=50$ in all experiments.

\paragraph{Effect of the smoothing scale $\sigma$.}
Next, we analyze the influence of the smoothing scale by fixing the number of samples to $N=10$ and varying $\sigma \in \{0.001, 0.01, 0.1, 0.2, 0.5\}$.
The results in \Cref{fig:ant_ablation_sigma} show that excessively small values of $\sigma$ lead to a significant performance degradation.
In particular, when $\sigma = 0.001$, the performance collapses, as the smoothing effect becomes negligible. On the other hand, overly large values of $\sigma$ (e.g., $\sigma = 0.5$) introduce excessive bias in the smoothed value function, leading to suboptimal performance.
Empirically, $\sigma = 0.2$ provides the best performance, striking a balance between stabilizing the learning dynamics and controlling approximation bias.
Similarly, although the optimal $\sigma$ may depend on the specific environment, $\sigma=0.2$ consistently provides stable and effective performance across tasks.
Accordingly, we use $\sigma = 0.2$ in all experiments.

\newpage
\subsection{Discrete reward experiment setting}\label{sec:experiment_detail}

In this section, we provide a detailed description of the discrete reward settings used in the main experiments.
Specifically, we first describe the overall experimental setup, including the hyperparameter configurations used for both DDPG and Soft DDPG.
We then present the construction of discrete reward environments, where standard dense reward functions are replaced with discontinuous and quantized reward signals.
These environments are designed to explicitly violate the smoothness assumptions underlying deterministic policy gradients, thereby providing a controlled setting to evaluate the robustness of the proposed method.
All experiments, including those presented in the main paper and appendix, were conducted on a single workstation equipped with an Intel Core i9-10900K CPU, an NVIDIA GeForce RTX 3080 Ti GPU with 12 GB memory, and 32 GB RAM. Each experiment was run on a single GPU. We begin by summarizing the hyperparameter settings used in all experiments.
Unless otherwise specified, these configurations are shared across both continuous and discrete reward environments to ensure a fair comparison.

\begin{table}[htbp]
\centering
\caption{Hyperparameter settings for the experiments}
\label{tab:hyperparameters}
\begin{tabular}{lc}
\toprule
\textbf{Hyperparameter} & \textbf{Value} \\
\midrule
$\sigma$ (Smoothing noise) & $0.2$ \\
$N$ (Number of smooth samples) & $50$ \\
Discount factor ($\gamma$) & $0.99$ \\
Batch size & $256$ \\
Target update rate ($\tau$) & $0.005$ \\
Actor learning rate & $10^{-4}$ \\
Critic learning rate & $10^{-4}$ \\
Replay buffer size & $1,000,000$ \\
Hidden units per layer & $400, 300$ \\
Number of hidden layers & $2$ \\
Total timesteps & $1,000,000$ \\
\bottomrule
\end{tabular}
\end{table}
Next, we describe the construction of the discrete reward functions for each environment.
In all cases, the original dense reward components are removed and replaced with discontinuous, task-specific reward signals designed to induce non-smooth learning dynamics.
\paragraph{Ant discrete reward setup}
In the modified Ant environment, the original dense reward components, such as forward-progress shaping and survival incentives, are removed and replaced with a sparse milestone-based reward structure along the forward direction. The modifications are as follows:
\begin{itemize}
    \item \textbf{Milestone-based distance rewards:} The agent receives a discrete bonus only when its horizontal position along the x-axis exceeds predefined distance thresholds for the first time. The milestone positions are set to $x \in \{0.5, 1.5, 3.0, 5.0, 10.0, 20.0\}$, and each milestone grants a reward of $50.0$.
    \item \textbf{Sparse curriculum structure:} The milestones are placed more densely in the early stage and more sparsely at longer distances. This design provides an initial learning signal even for short forward movements, while still encouraging the agent to achieve sustained long-horizon locomotion.
    \item \textbf{Flip penalty:} A terminal penalty of $-10.0$ is applied when the episode ends due to failure, such as the ant flipping over, excluding time-limit truncation. This discourages unstable postures during exploration.
    \item \textbf{Survival reward removal:} The standard per-step survival reward is set to zero. This prevents the agent from exploiting a local optimum in which it simply remains alive without making meaningful forward progress.
\end{itemize}

\paragraph{HalfCheetah discrete reward setup}
In the modified HalfCheetah environment, the original dense reward composed of forward velocity and control-related shaping is replaced with a discretized velocity-based reward. The modifications are as follows:
\begin{itemize}
    \item \textbf{Velocity-binned reward:} The agent's forward velocity $v_x$ is partitioned into discrete intervals defined by the thresholds $\{0.0, 1.0, 3.0, 5.0\}$. The reward is assigned according to the velocity bin in which the agent currently lies.
    \item \textbf{Reward assignments:} The step reward is given as
    \begin{itemize}
        \item $0.0$ if $v_x \le 0.0$,
        \item $0.5$ if $0.0 < v_x \le 1.0$,
        \item $1.0$ if $1.0 < v_x \le 3.0$,
        \item $2.0$ if $3.0 < v_x \le 5.0$,
        \item $3.0$ if $v_x > 5.0$.
    \end{itemize}
    \item \textbf{Quantized performance signal:} As the forward speed increases, the reward changes only when the velocity crosses one of the bin boundaries, thereby creating a piecewise-constant and discontinuous learning signal.
    \item \textbf{Dense shaping removal:} The original dense shaping terms are discarded so that the learning behavior is driven entirely by the discretized velocity levels.
\end{itemize}

\paragraph{Hopper discrete reward setup}
In the modified Hopper environment, the original dense reward is replaced with a sparse milestone-based reward defined on the agent's forward displacement. The modifications are as follows:
\begin{itemize}
    \item \textbf{Forward milestone reward:} The agent receives a discrete reward whenever its horizontal position increases by more than a predefined threshold relative to the last rewarded position. Specifically, if the forward displacement exceeds $\Delta x = 0.5$, a reward of $5.0$ is granted for each crossed milestone.
    \item \textbf{Accumulated milestone counting:} If the agent advances by multiple thresholds within a single step, the reward is scaled proportionally to the number of crossed milestones. This preserves the cumulative forward-progress information while maintaining a discretized signal.
    \item \textbf{Minimal survival reward:} A very small survival reward is provided at each non-terminal step. This weak auxiliary signal prevents the agent from immediately collapsing while ensuring that the primary learning signal still comes from forward milestone crossings.
    \item \textbf{Dense reward removal:} All original continuous reward components, including forward-velocity shaping and control-related terms, are discarded so that the agent is trained entirely under the proposed discrete reward structure.
\end{itemize}

\paragraph{Walker2d discrete reward setup}
In the modified Walker2d environment, the original dense reward is replaced with a sparse milestone-based forward-progress reward. The modifications are as follows:
\begin{itemize}
    \item \textbf{Forward milestone reward:} The agent receives a discrete reward only when its horizontal displacement from the last rewarded position exceeds a threshold of $\Delta x = 0.2$. Each crossed milestone yields a reward of $2.0$.
    \item \textbf{Accumulated milestone counting:} If multiple displacement thresholds are crossed within a single step, the reward is scaled proportionally to the number of crossed milestones. This preserves coarse forward-progress information while maintaining a discontinuous reward signal.
    \item \textbf{Minimal survival reward:} A very small constant reward is added at each step. This prevents immediate degeneration of behavior while ensuring that the dominant learning signal still comes from milestone-based forward progress.
    \item \textbf{Discontinuity emphasis:} Between milestone crossings, the forward reward remains zero, and reward is generated only when the agent passes the threshold. This creates a highly discontinuous signal and makes the task substantially harder than the standard dense-reward setting.
    \item \textbf{Dense reward removal:} The original dense reward components are discarded so that the agent is trained entirely under the proposed discrete locomotion reward structure.
\end{itemize}

\paragraph{Humanoid discrete reward setup}
In the modified Humanoid environment, the original dense reward is replaced with an extremely sparse and highly discretized reward function defined by posture and velocity conditions. The modifications are as follows:
\begin{itemize}
    \item \textbf{Posture constraint:} Rewards are provided only when the humanoid maintains an upright and stable posture. Specifically, the torso height must satisfy $z > 1.18$, and the lateral velocity magnitude must remain below $0.18$.
    \item \textbf{Low-speed reward:} If the posture constraint is satisfied and the forward velocity is smaller than $0.12$, the agent receives a reward of $3.0$. This provides a small discrete signal for stable but slow locomotion.
    \item \textbf{High-speed reward:} If the posture constraint is satisfied and the forward velocity exceeds $1.6$, the agent becomes eligible for a larger reward. However, the reward of $20.0$ is granted only after this high-speed condition is maintained for at least six consecutive steps.
    \item \textbf{Persistence requirement:} The consecutive-step counter is reset whenever the posture constraint is violated or the forward velocity leaves the high-speed regime. This makes the large reward dependent on sustained stable locomotion rather than transient bursts of speed.
    \item \textbf{Dense reward removal:} All original dense reward components are discarded so that the agent is trained solely under the proposed discrete reward mechanism.
\end{itemize}

\paragraph{InvertedDoublePendulum discrete reward setup}
In the modified InvertedDoublePendulum environment, the original dense reward is replaced with a discrete reward structure based on angular stability and long-term survival. The modifications are as follows:
\begin{itemize}
    \item \textbf{Step-wise angular reward:} Let $\theta_1$ and $\theta_2$ denote the two pendulum angles, and define $m = \max(|\theta_1|, |\theta_2|)$. At each step, the agent receives a discrete reward according to the angular region:
    \begin{itemize}
        \item $20.0$ if $m < 0.05$,
        \item $5.0$ if $0.05 \le m < 0.15$,
        \item $1.0$ if $0.15 \le m < 0.4$,
        \item $0.0$ otherwise.
    \end{itemize}
    This creates a staircase-shaped reward landscape in which more accurate balancing yields a larger discrete reward.
    \item \textbf{Survival milestone bonus:} Additional sparse bonuses are given when the agent survives for a sufficiently long duration. Specifically, rewards of $500.0$ are granted once when the step count reaches 500 and 1000, respectively.
    \item \textbf{Long-horizon credit assignment:} The survival milestones introduce large delayed rewards, making the task suitable for testing whether the algorithm can propagate long-term value information under sparse feedback.
    \item \textbf{Failure penalty:} If the episode terminates due to failure, excluding time-limit truncation, the reward is overwritten to $-10.0$. This penalizes collapse while keeping the penalty moderate enough to avoid completely suppressing exploration.
    \item \textbf{Dense reward removal:} All original dense reward components are discarded so that the agent is trained entirely under the proposed discrete balancing reward structure.
\end{itemize}

\paragraph{InvertedPendulum discrete reward setup}
In the modified InvertedPendulum environment, the original dense reward is replaced with a highly sparse reward structure based on angular stability and long-term survival. The modifications are as follows:
\begin{itemize}
    \item \textbf{Angular reward:} Let $\theta$ denote the pendulum angle from the upright position. At each step, the agent receives a reward of $10.0$ only when $|\theta| < 0.02$, and receives $0.0$ otherwise. This creates a narrow success region in which reward is available only for highly accurate balancing.
    \item \textbf{Survival milestone bonus:} Additional sparse bonuses are granted when the agent survives for a sufficiently long duration. Specifically, rewards of $200.0$ are given once when the step count reaches 500 and 1000, respectively.
    \item \textbf{Long-horizon credit assignment:} The survival milestones provide delayed rewards that encourage sustained stabilization, making the task suitable for testing long-term value propagation under sparse feedback.
    \item \textbf{Failure penalty:} If the episode terminates due to failure, excluding time-limit truncation, the reward is overwritten to $-20.0$. This discourages immediate collapse and distinguishes catastrophic failure from ordinary zero-reward states.
    \item \textbf{Dense reward removal:} All original dense reward components are discarded so that the agent is trained entirely under the proposed discrete balancing reward structure.
\end{itemize}

\section{Additional discrete reward experiment}\label{app:discrete_additional}

To further validate the robustness of the proposed method, we conduct additional experiments on a diverse set of Gym environments \citep{brockman2016openai}, including BipedalWalker, LunarLander, Pendulum, and MountainCar. 
These environments differ significantly in their dynamics and control objectives, providing a broader evaluation beyond the benchmarks presented in the main text.

All experiments follow the same training protocol and hyperparameter settings as described in \Cref{sec:experiment_detail}, ensuring consistency and fair comparison. 
For each environment, we construct discrete reward variants by replacing the original dense reward functions with quantized or sparse reward signals, following the same design principle used in the main experiments.

The quantitative results are summarized in \Cref{tab:additional_result}.
Consistent with the main experiments, Soft DDPG demonstrates clear advantages in the discrete reward settings across most environments.
In particular, it outperforms the baseline in BipedalWalker, MountainCar, and Pendulum, while achieving comparable performance in LunarLander.

These results further support our main claim that Soft DDPG is particularly effective in environments with non-smooth or quantized reward structures, where the action-gradient of the critic becomes unreliable.
By leveraging the $\sigma$-smoothed Bellman formulation, the proposed method maintains stable and informative policy updates, leading to improved performance under such challenging conditions.

In contrast, under continuous reward settings, vanilla DDPG generally achieves better performance.
This behavior is expected, as the smoothing mechanism introduces bias when the underlying value function is already smooth and well-behaved. 

Overall, these results demonstrate Soft DDPG  provides substantial gains in robustness and stability when the reward landscape is discontinuous or non-differentiable.

\begin{figure*}[t]
    \centering
    \subfloat[Bipedalwalker (Cont.)]{%
        \includegraphics[width=0.23\textwidth]{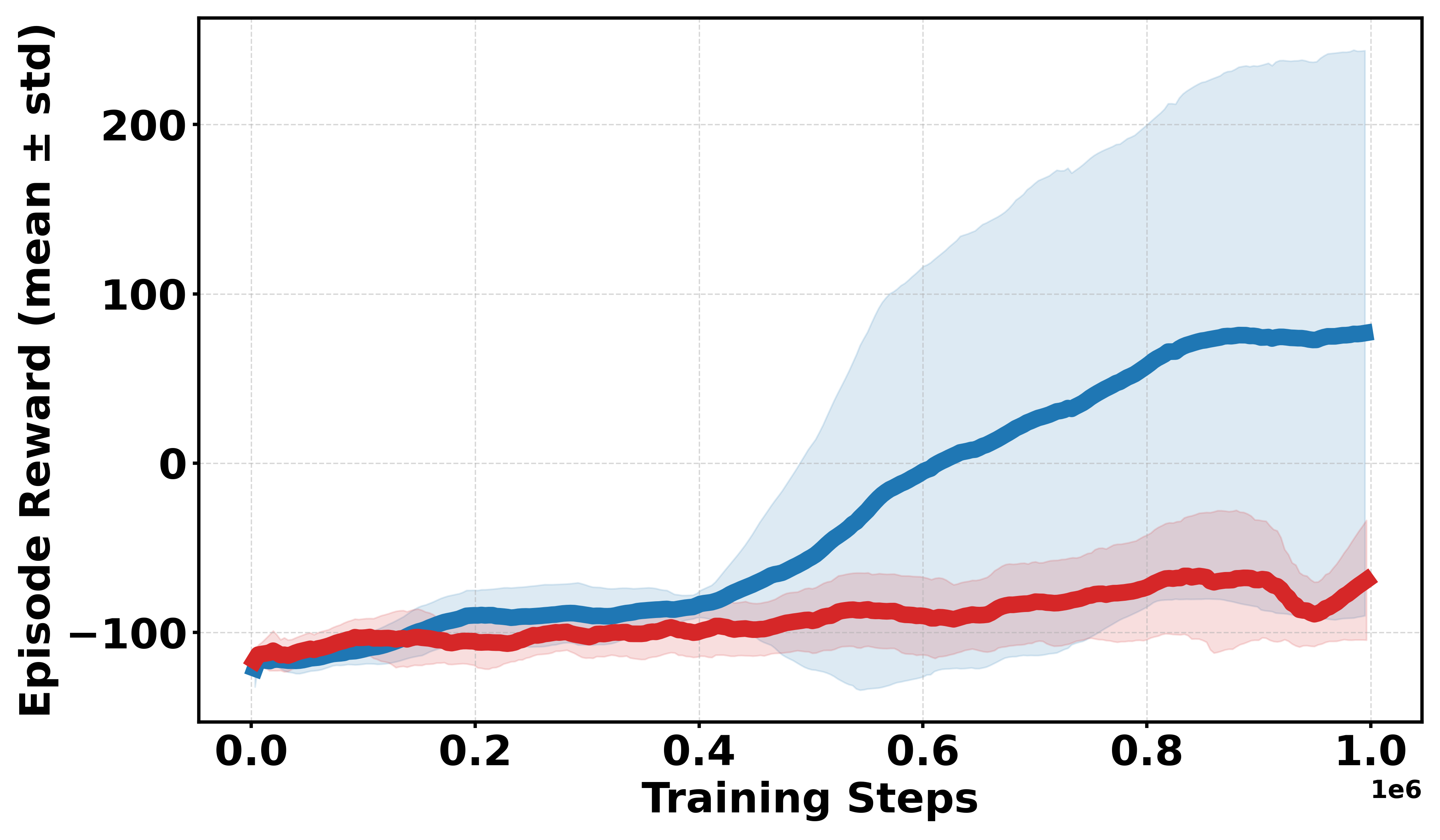}%
    }\hfill
    \subfloat[Lunarlander (Cont.)]{%
        \includegraphics[width=0.23\textwidth]{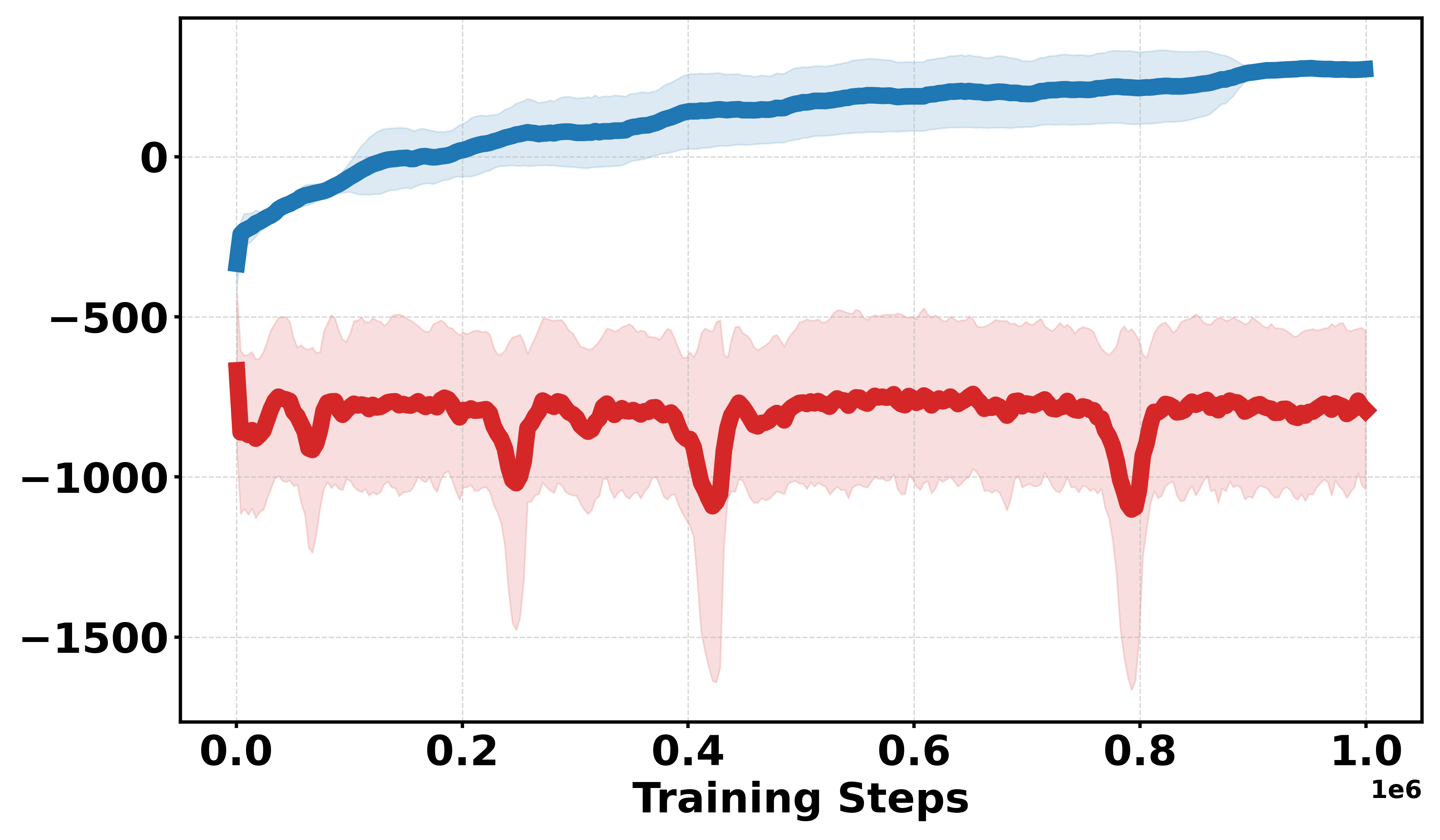}%
    }\hfill
    \subfloat[Pendulum (Cont.)]{%
        \includegraphics[width=0.23\textwidth]{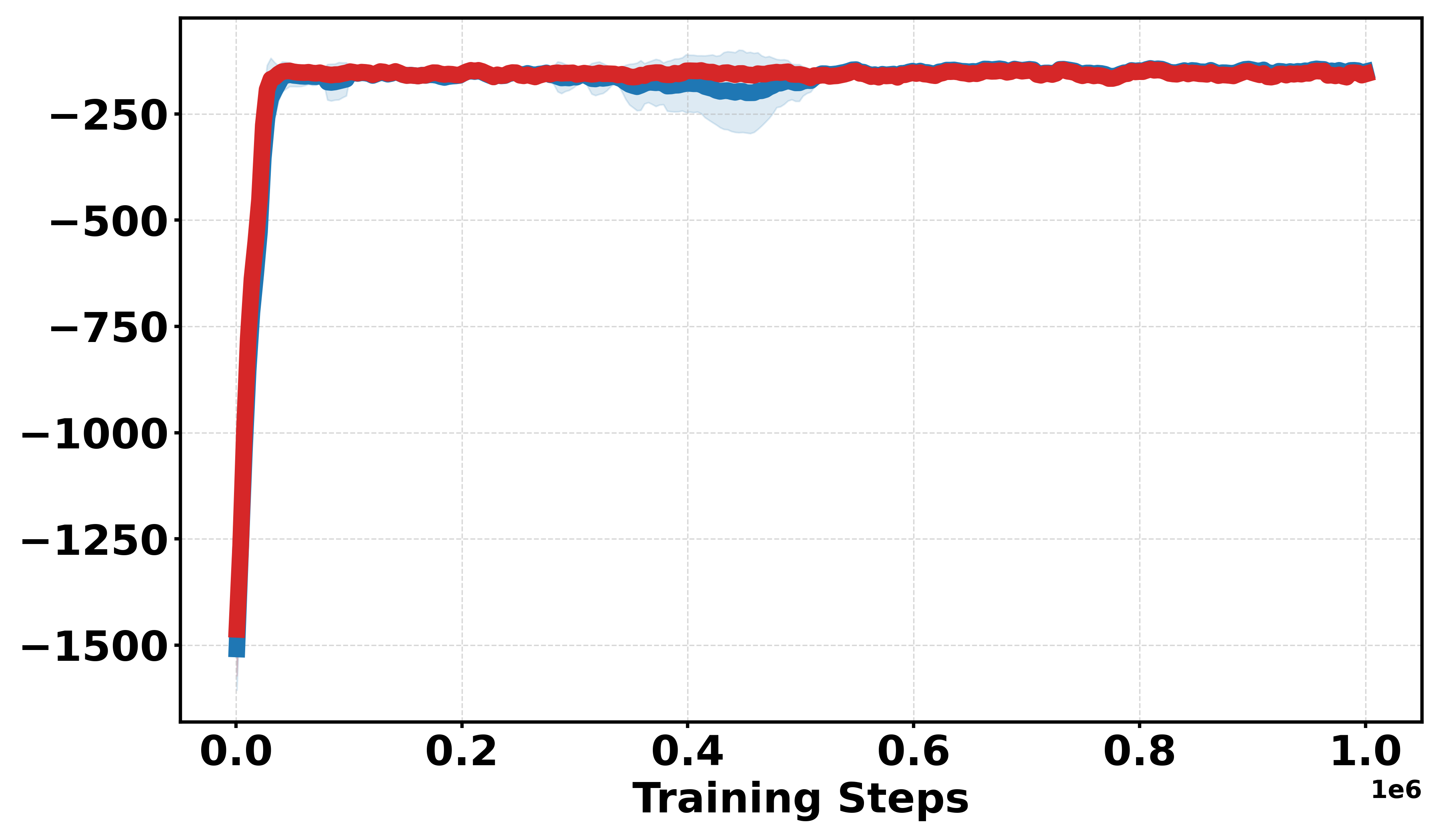}%
    }\hfill
    \subfloat[Mountaincar (Cont.)]{%
        \includegraphics[width=0.23\textwidth]{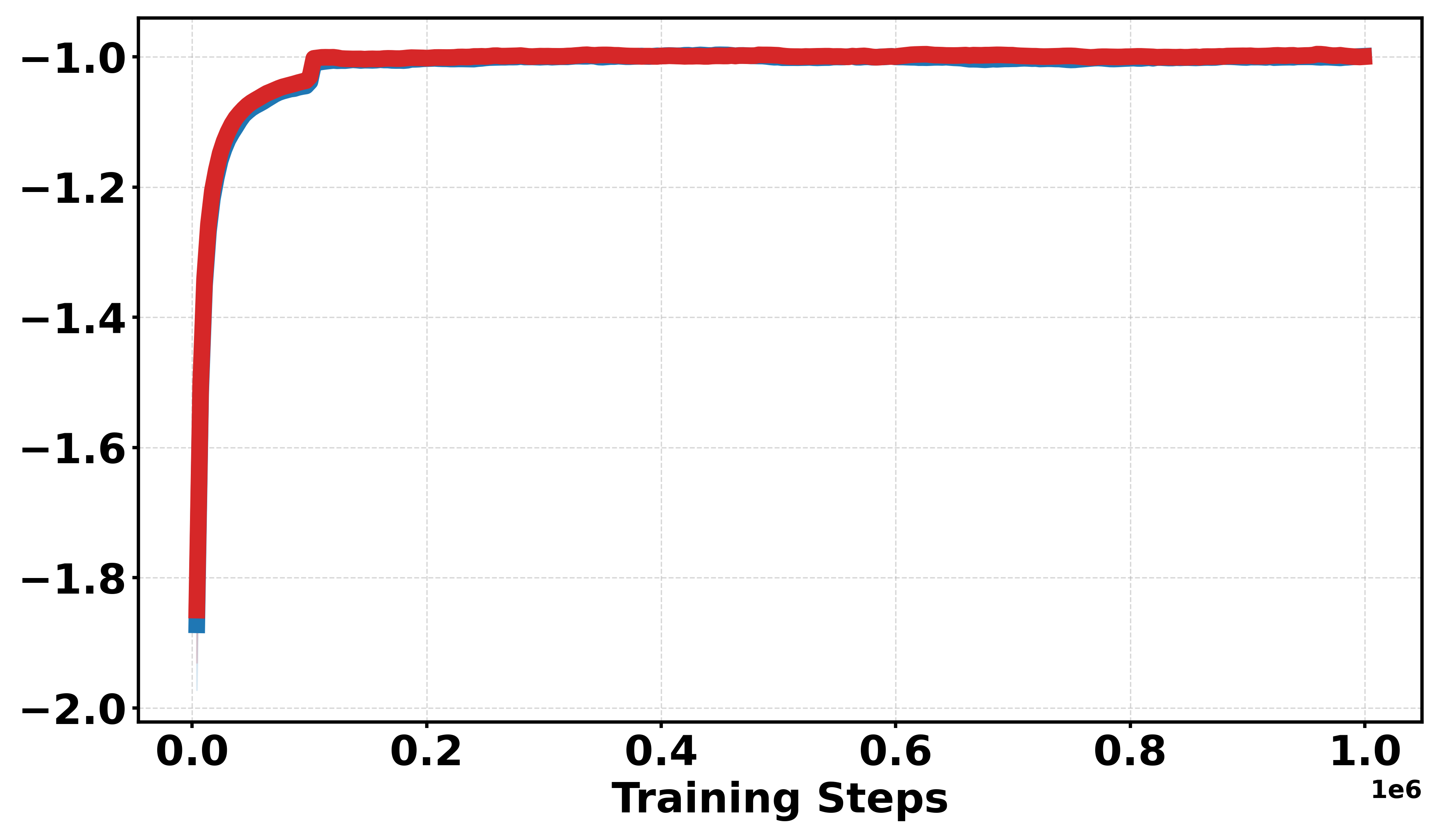}%
    }

    \vspace{0.5em} 

    \subfloat[Bipedalwalker (Disc.)]{%
        \includegraphics[width=0.23\textwidth]{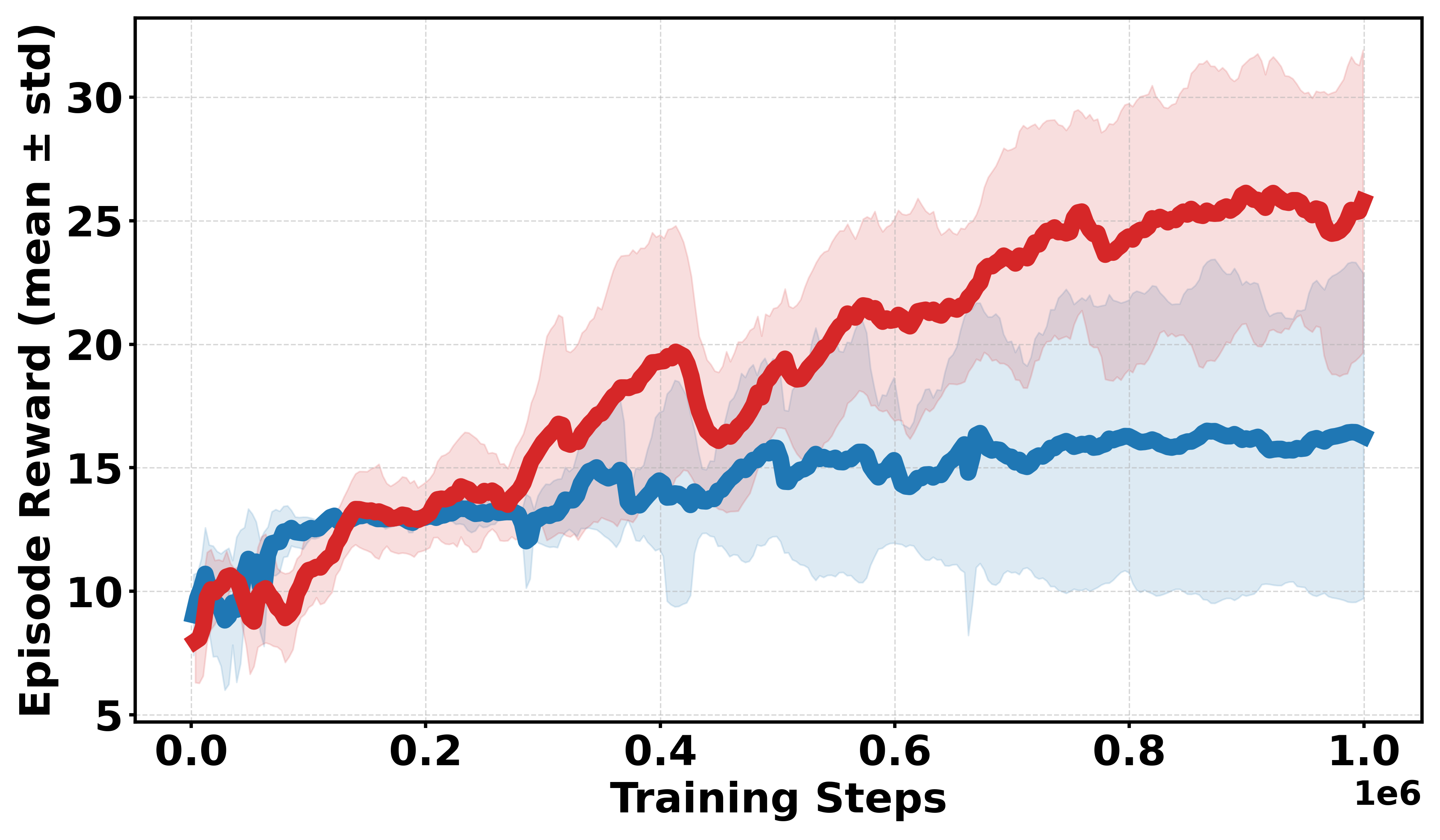}%
    }\hfill
    \subfloat[Lunarlander (Disc.)]{%
        \includegraphics[width=0.23\textwidth]{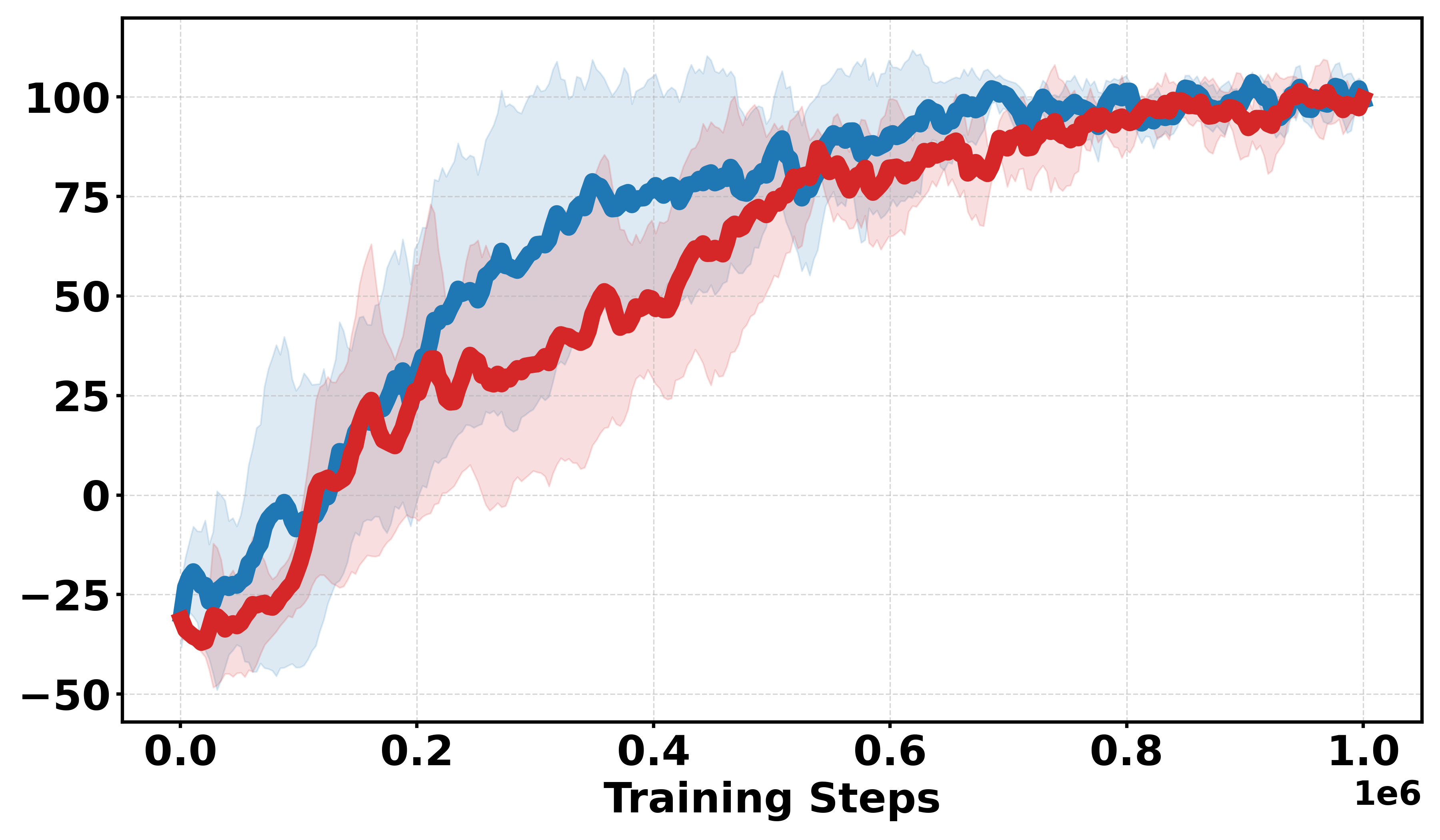}%
    }\hfill
    \subfloat[Pendulum (Disc.)]{%
        \includegraphics[width=0.23\textwidth]{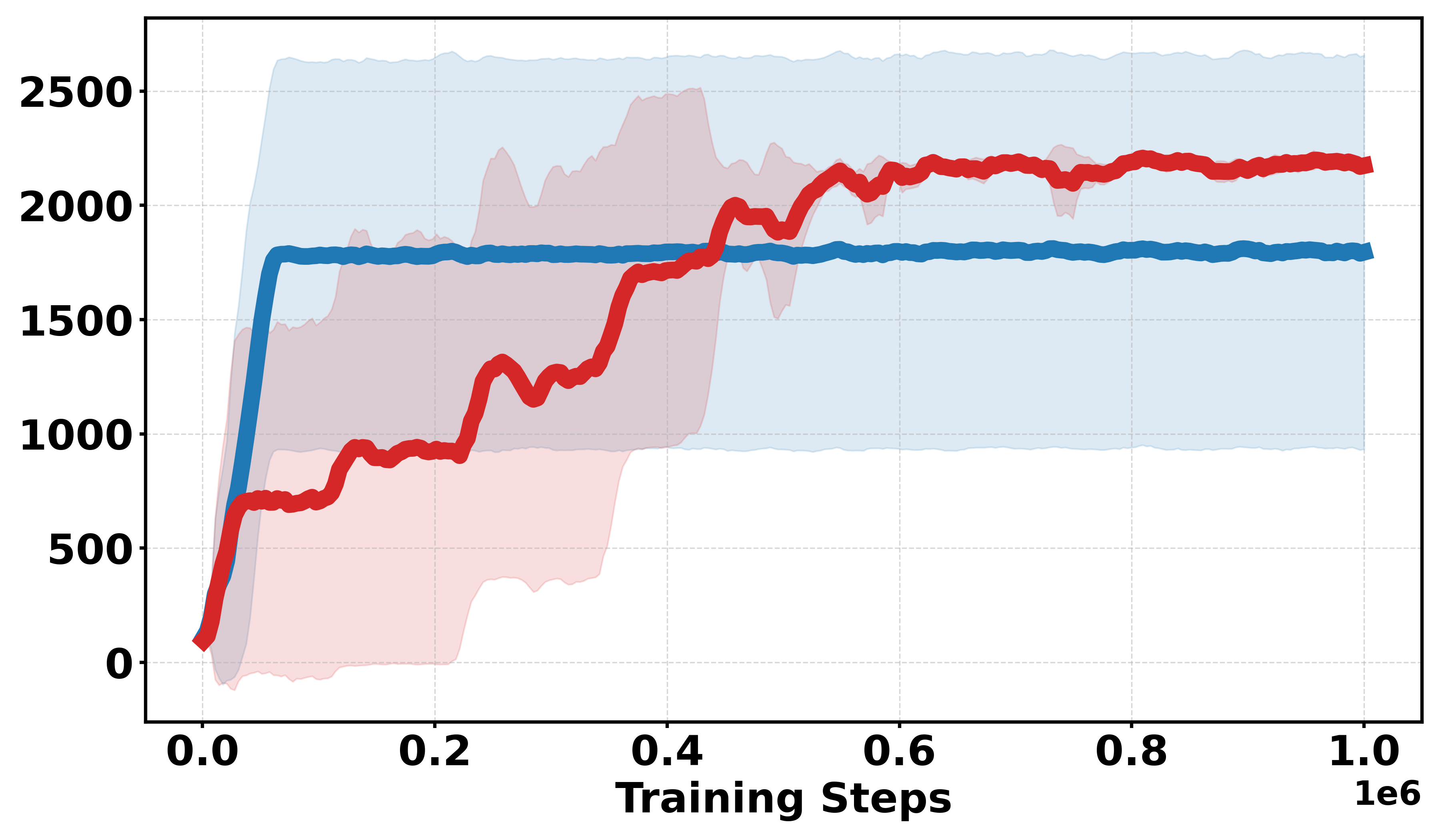}%
    }\hfill
    \subfloat[Mountaincar (Disc.)]{%
        \includegraphics[width=0.23\textwidth]{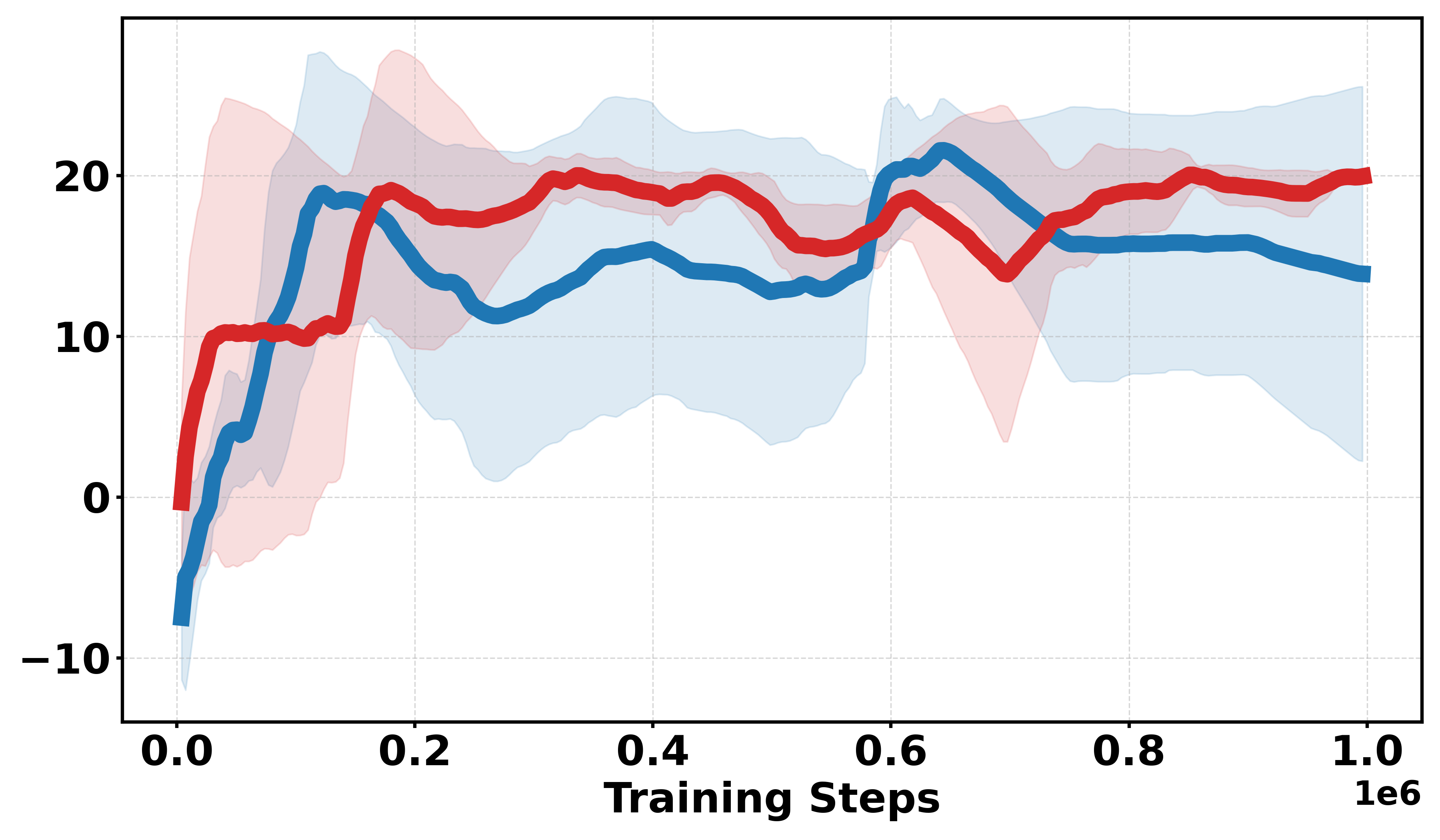}%
    }

    \vspace{0cm} 

    \includegraphics[width=0.6\textwidth]{figures/legend_horizontal.png}

    \vspace{-0cm} 

    \caption{Learning curves on various benchmarks: Continuous (top) vs. Discrete (bottom) control. The solid lines represent the mean and the shaded areas represent the standard deviation across multiple seeds.}
    \label{fig:main_benchmarks_comparison}
\end{figure*}

\begin{table}[h!]
\centering
\footnotesize
\setlength{\tabcolsep}{4pt}
\caption{Performance comparison in discrete- and continuous-reward environments (mean $\pm$ std over 5 seeds). Bold values indicate the better result within each reward setting.}
\label{tab:combined_results_gym}
\begin{tabular}{lcccc}
\toprule
\multirow{2}{*}{Environment}
& \multicolumn{2}{c}{Discrete}
& \multicolumn{2}{c}{Continuous} \\
\cmidrule(lr){2-3} \cmidrule(lr){4-5}
& DDPG & Soft DDPG
& DDPG & Soft DDPG\\
\midrule
Bipedalwalker
& $16.13 \pm 6.22$ & $\mathbf{25.42 \pm 5.46}$
& $\mathbf{76.95 \pm 164.00}$ & $-80.48 \pm 23.73$ \\

Lunarlander
& $\mathbf{99.03 \pm 3.13}$ & $98.56 \pm 5.17$
& $\mathbf{273.11 \pm 3.77}$ & $-787.54 \pm 252.02$ \\

Pendulum
& $1797.67 \pm 860.55$ & $\mathbf{2182.19 \pm 19.22}$
& $\mathbf{-149.60 \pm 2.84}$ & $-156.72 \pm 3.28$ \\

Mountaincar
& $15.06 \pm 9.40$ & $\mathbf{19.36 \pm 0.93}$
& $\mathbf{-1.00 \pm 0.00}$ & $\mathbf{-1.00 \pm 0.00}$ \\

\bottomrule
\end{tabular}
\label{tab:additional_result}
\end{table}

The detailed discrete reward formulations for each environment are described below.

\paragraph{BipedalWalker discrete reward setup}
In the modified BipedalWalker environment, the original dense reward is replaced with a sparse distance-based milestone reward together with a discrete terminal fall penalty. The modifications are as follows:
\begin{itemize}
    \item \textbf{Distance milestone reward:} The agent receives a discrete reward when its forward position crosses a predefined distance threshold for the first time during an episode. The thresholds are set to $x \in \{0.5, 1.0, 5.0, 10.0\}$ with corresponding rewards $\{1.0, 2.0, 10.0, 20.0\}$.
    \item \textbf{One-time reward per threshold:} Each milestone reward is granted only once per episode, based on the maximum forward distance achieved so far. This ensures that the reward reflects genuine progress rather than repeated oscillations around the same position.
    \item \textbf{Fall penalty:} If the episode terminates due to failure rather than time-limit truncation, the agent receives a terminal penalty of $-10.0$. This penalizes collapse and encourages stable walking behavior.
    \item \textbf{Dense reward removal:} All original dense reward components, including forward-progress shaping and torque-related penalties, are discarded so that the agent is trained entirely under the proposed discrete milestone structure.
\end{itemize}

\paragraph{LunarLander discrete reward setup}
In the modified LunarLander environment, the original dense reward is replaced with a staged discrete reward structure that reflects the sequential procedure of successful landing. The modifications are as follows:
\begin{itemize}
    \item \textbf{Approach reward:} A reward of $5.0$ is granted once when the lander enters the central horizontal region of the landing pad, i.e., when $|x| < 0.2$.
    \item \textbf{Hover reward:} After the approach milestone has been achieved, an additional reward of $10.0$ is granted once when the lander descends below a low-altitude threshold, i.e., when $y < 0.3$.
    \item \textbf{Success reward:} If the episode terminates with both legs in contact with the ground and the lander remains within the landing zone ($|x| < 0.2$), a terminal reward of $100.0$ is assigned.
    \item \textbf{Crash penalty:} If the episode terminates without both legs contacting the ground, excluding time-limit truncation, a terminal penalty of $-50.0$ is applied.
    \item \textbf{One-time milestone reward:} The approach and hover rewards are granted only once per episode, preventing repeated reward collection from oscillatory motion around the same region.
    \item \textbf{Dense reward removal:} All original dense reward components, such as fuel-related penalties and continuous shaping terms, are discarded so that the agent is trained entirely under the proposed discrete landing reward structure.
\end{itemize}

\paragraph{Pendulum discrete reward setup}
In the modified Pendulum environment, the original dense reward is replaced with a  discrete reward structure designed to encourage swing-up, near-upright stabilization, and sustained balancing. The modifications are as follows:
\begin{itemize}
    \item \textbf{Angle-band reward:} Let $\theta$ denote the pendulum angle from the upright position. A discrete reward is assigned according to the angular deviation:
    \begin{itemize}
        \item $8.0$ if $|\theta| < 0.10$,
        \item $4.0$ if $0.10 \le |\theta| < 0.25$,
        \item $2.0$ if $0.25 \le |\theta| < 0.50$,
        \item $0.5$ if $0.50 \le |\theta| < 1.00$,
        \item $0.0$ otherwise.
    \end{itemize}
    This provides a staircase-shaped reward signal that guides the swing-up process toward the upright region.
    \item \textbf{Near-upright stabilization bonus:} When the pendulum is sufficiently close to upright ($|\theta| < 0.25$), an additional reward is given based on the angular velocity magnitude $|\dot{\theta}|$:
    \begin{itemize}
        \item $4.0$ if $|\dot{\theta}| < 0.5$,
        \item $2.0$ if $0.5 \le |\dot{\theta}| < 1.0$.
    \end{itemize}
    This encourages not only reaching the upright region but also reducing oscillatory motion around it.
    \item \textbf{Sustained hold bonus:} If the pendulum remains within a near-upright and low-velocity region defined by $|\theta| < 0.15$ and $|\dot{\theta}| < 0.7$, a hold counter is accumulated. Additional bonuses are granted when this condition is maintained for 10, 30, and 60 consecutive steps, with rewards of $20.0$, $40.0$, and $80.0$, respectively.
    \item \textbf{Hierarchical discrete feedback:} The resulting reward structure combines coarse swing-up guidance, local stabilization incentives, and delayed persistence rewards, thereby forming a hierarchical discrete control objective.
    \item \textbf{Dense reward removal:} All original dense reward components are discarded so that the agent is trained entirely under the proposed discrete reward mechanism.
\end{itemize}

\paragraph{MountainCar discrete reward setup}
In the modified MountainCar environment, the original dense reward is replaced with a staged discrete reward structure that reflects the characteristic swing-up behavior required to reach the goal. The modifications are as follows:
\begin{itemize}
    \item \textbf{Backward swing reward:} A reward of $10.0$ is granted once when the car first reaches the backward position threshold $x < -0.7$. This encourages the agent to perform the necessary backswing motion to build momentum.
    \item \textbf{Forward milestone reward:} A reward of $20.0$ is granted once when the car first reaches the forward position threshold $x > -0.1$. This provides an intermediate sparse signal for successful forward progress after the backswing phase.
    \item \textbf{Goal reward:} If the car reaches the goal position $x \ge 0.5$, an additional reward of $100.0$ is assigned and the episode is terminated.
    \item \textbf{One-time milestone reward:} The backward-swing and forward-progress rewards are granted only once per episode, preventing repeated reward collection from oscillation around the same thresholds.
    \item \textbf{Per-step penalty:} A small step penalty of $-0.01$ is applied at every step. This weakly encourages faster completion while keeping the main learning signal dominated by the discrete milestones.
    \item \textbf{Dense reward removal:} All original dense reward components are discarded so that the task is learned entirely from the proposed sparse staged reward structure.
\end{itemize}


\newpage
\newpage

\end{document}